\documentclass[10pt]{article}

\usepackage{graphicx}
\usepackage{color}

\usepackage[authoryear]{natbib}

\usepackage{bm}
\usepackage{amsmath}
\usepackage{amsfonts}
\usepackage{amsthm}
\usepackage{amssymb}
\usepackage{mathtools}
\mathtoolsset{showonlyrefs=true}

\DeclareMathOperator{\sign}{sign}

\DeclareMathOperator*{\argmax}{arg~max}

\usepackage{url}
\usepackage{subcaption}
\usepackage{algorithm}
\usepackage{algorithmic}

\makeatletter
\renewcommand{\ALG@name}{Procedure}
\makeatother

\usepackage[hidelinks]{hyperref}

\title{Selective Inference for Changepoint detection by Recurrent Neural Network}

\date{\today}

\makeatletter
\def\@fnsymbol#1{\ensuremath{\ifcase#1\or
{1}\or
{2}\or
{3}\or
{\dagger}\or
\else\@ctrerr\fi}}
\makeatother

\author{
Tomohiro Shiraishi\thanks{Nagoya University} ,
Daiki Miwa\thanks{Nagoya Institute of Technology} ,\\
Vo Nguyen Le Duy\thanks{RIKEN} ,
Ichiro Takeuchi\footnotemark[1] \footnotemark[3] \thanks{Corresponding author. e-mail: ichiro.takeuchi@mae.nagoya-u.ac.jp}
}
\begin{document}

\maketitle

\thispagestyle{empty}

\begin{abstract}
    \noindent
    In this study, we investigate the quantification of the statistical reliability of detected change points (CPs) in time series using a Recurrent Neural Network (RNN).
Thanks to its flexibility, RNN holds the potential to effectively identify CPs in time series characterized by complex dynamics.
However, there is an increased risk of erroneously detecting random noise fluctuations as CPs.
The primary goal of this study is to rigorously control the risk of false detections by providing theoretically valid p-values to the CPs detected by RNN.
To achieve this, we introduce a novel method based on the framework of Selective Inference (SI).
SI enables valid inferences by conditioning on the event of hypothesis selection, thus mitigating selection bias.
In this study, we apply SI framework to RNN-based CP detection, where characterizing the complex process of RNN selecting CPs is our main technical challenge.
We demonstrate the validity and effectiveness of the proposed method through artificial and real data experiments.

\end{abstract}

\newpage
\section{Introduction}
\label{sec:introduction}
In this study, we are concerned with change-point (CP) detection in time-series by using a deep neural network (DNN).
CP detection is one of the most important problems in time-series analysis with various applications in a variety of fields such as finance~\citep{pepelyshev2016real,hsu1982bayesian}, biology~\citep{wang2011nonparametric,takeuchi2009potential}, and engineering~\citep{basu2007automatic,shi2021dual}.
Identifying abrupt shifts in time-series data plays a crucial role in system maintenance and anomaly detection, as it is essential for making timely interventions.
While traditional statistical methods have been employed for decades to tackle this problem, they often struggle to adapt to the intricate patterns and complexities present in real-world data.
The recent advances of DNNs has offered a promising alternative approach.
Leveraging their capacity to automatically extract complex patterns from data, DNNs have shown remarkable potential in enhancing the accuracy of CP detection.

However, with complex models such as DNNs, ensuring the reliability of detected CPs is more challenging compared to the simple models traditionally used in the literature of statistics.
Especially, when making high-stakes decisions based on detected CPs, it is essential to quantitatively evaluate the reliability of the CPs.
Due to their flexibility, DNNs can learn the dynamics of complex signals but are at a higher risk of mistakenly detecting noise fluctuations as CPs.
The goal of this study is to quantitatively assess the statistical reliability of CPs detected by DNN within the framework of statistical hypothesis testing.
Specifically, we introduce a novel method to calculate theoretically valid $p$-values for the CPs detected by DNN, thereby ensuring that the probability of false detection of these CPs remains below a certain threshold.

In time-series CP detection, it is often challenging to obtain data where actual changes have occurred.
Therefore, it is common to first train a model that predicts normal time-series without any CPs.
Then, discrepancies between the predictions made by the learned time-series model and the actual observed time-series are considered as indications of CPs.
As a simple yet effective DNN for time-series prediction, we employ the Recurrent Neural Network (RNN) in this study.
RNNs are a class of DNNs specifically designed for handling sequential data.
In a time-series prediction task, an RNN processes the data sequentially, step by step, where each input at a given time step is combined with the model's internal state, and the result is used to generate predictions for the next time step.
This recurrent nature allows RNNs to account for the evolving nature of time-series data, making them well-suited for time-series predictions.

The problem of calculating $p$-values for detected CPs in time-series is challenging, even when simple model and algorithm are used.
This is because the selection of CPs and their testing must be performed on the same time-series sequence.
Because CP detection algorithms inherently select points that seem likely to be CPs, using the same time-series sequence for testing these CPs results in \emph{selection bias}.
Therefore, in traditional statistical contexts, the assessment of the reliability of CPs is performed based on asymptotic theory~(e.g., refer to \citep{mika1999fisher,shuang2015mstatistic,page1954continuous,klaus2014multiscale}), which requires assumptions that the time-series sequence is sufficiently long and has minimal or no correlation.

Recently, a new approach utilizing \emph{selective inference (SI)} has successfully addressed the selection bias issue in testing detected CPs.
In SI, CP detection is treated as a hypothesis selection, and statistical inference \emph{conditional on the hypothesis selection event} is considered.
By conditioning on the hypothesis selection event (i.e., the event that the CP is selected in our problem), the influence of the fact that the CP is being selected from the same time-series sequence can be eliminated.
Unfortunately, the class of hypothesis selection events that can be handled by the current state of SI is limited to events that can be characterized as the intersection of linear and/or quadratic inequalities with respect to the data.
As a result, the aforementioned SI approach is only applicable to simple CP detection algorithms.
The primary contribution of this study is to extend the SI-based CP testing approach, which was previously limited to simple CP detection algorithms, to RNN-based CP detection algorithms.

\paragraph{Contributions}

Our contributions are summarized as follows:
\begin{itemize}
    \item  We introduce a novel setup of testing the RNN-driven CPs. Compared the literature of RNN-based CP detection, we present a new challenge in addressing the intricate selection process of RNNs to assess the reliability of CPs.
    \item We conquer the challenge by proposing a method for efficiently handling the recurrent structure of RNNs, based on the literature of SI. This allows us to conduct a valid statistical inference for the CPs detected by RNNs. To our knowledge, this is the first method that can provide a valid $p$-value for the RNN-driven CPs.
    \item We conduct experiments on both synthetic and real-world datasets to demonstrate the superior performance of the proposed method compared to the existing studies.
\end{itemize}

\paragraph{Related Works}

Existing studies on CP detection using DNNs have shown promising results in a variety of applications~\citep{ahad2021semi,deldari2021time,ryck2021change,du2022finder}.
These studies leverage the capabilities of DNNs to automatically extract complex patterns from time-series data, making them well-suited for identifying abrupt shifts.
In time-series CP detection, obtaining data with actual changes is often difficult, so it is common to train a model for predicting normal time-series, and discrepancies between the model predictions and the observed time-series are considered as indications of CPs~\citep{yaacob2010arima,yu2016improved,mohsin2019deepant}.
Among various types of DNNs, RNNs are useful for CP detection problems due to their inherent ability to capture temporal dependencies and model sequential data effectively~\citep{malhotra2016lstm,chung2016hierarchical,chang2018kernel}.

The challenge of quantitatively assessing the reliability of DNNs, not only in the context of CP detection but more broadly, remains an open problem.
For example, attempts have been made to quantify the uncertainty of DNNs through Bayesian modeling~\citep{gal2015bayesian,auld2007bayesian,shawe2011practical}.
While Bayesian DNNs enables an understanding of prediction uncertainty and reliability, it may not be suitable for precisely controlling the false detection rates of CPs, as we aim to do in this study.
Furthermore, while the prediction performance of DNN used for CP detection problems can be evaluated using techniques such as cross-validation, it is important to note that this does not directly measure the reliability of the detected CPs.
Moreover, even when multiple time-series sequences containing known CPs are available and evaluations of CP detection algorithms are possible, it does not necessarily imply the reliability of detected specific CPs in a particular time-series sequence.
To the best of our knowledge, there is currently no method for rigorously quantifying the statistical reliability of the CPs detected by DNNs.

SI was first introduced within the context of reliability evaluation for linear model features when they were selected using a feature selection algorithm, such as Lasso~\citep{lee2016exact}, and extended to various directions, e.g., \citep{tibshirani2016exact,suzumura2017selective,terada2017selective,liu2018more,tian2018selective,yamada2018post,tanizaki2020computing,benjamini2020selective,panigrahi2022exact,neufeld2022tree,duy2022more,miwa2023valid}.
The fundamental idea of SI is to perform an inference conditional on the event of selecting a hypothesis, which mitigates the selection bias issue even when the hypothesis is selected and tested using the same data.
To conduct SI, it is necessary to derive the sampling distribution of test statistics conditional on the hypothesis selection event.
Currently, a widely known SI approach, which we call \emph{polyhedron-based SI}, is applicable only when the hypothesis selection event can be characterized as an intersection of linear and/or quadratic inequalities.
Therefore, it is challenging to apply polyhedron-based SI to complex algorithms like RNNs.

The framework for conducting reliability evaluations of detected CPs using SI was first proposed by \citep{hyun2016exact} in the context of fused LASSO-based CP detection.
Subsequently, the framework was extended to various different CP detection methods~\citep{umezu2017selective, hyun2018post, duy2020computing, sugiyama2021valid, jewell2022testing}, but not to CP detection algorithms using DNNs.
To the best of our knowledge, SI was first applied to inference on DNNs in \citep{duy2022quantifying} for image segmentation problems, and it was extended to evaluate the saliency maps in \citep{miwa2023valid}.
These studies have focused on Convolutional Neural Networks (CNNs) and have introduced a computational method called \emph{Parametric Programming (PP)} to make it possible to condition more complex hypothesis selection events.
In this study, we also adopt PP to compute conditional sampling distribution when the hypotheses (CPs) are selected by RNNs.
Unlike CNNs, RNNs have a recurrent structure, making the extension from CNNs to RNNs non-trivial.
Moreover, the above two studies have focused on computer vision tasks, which are fundamentally different from our time-series CP detection task.
For reproducibility, our implementation is available at \url{https://github.com/shirara1016/si_for_cpd_by_rnn}

\newpage
\section{Problem Setup}
\label{sec:sec2}
In this paper, we consider two types of CP detection problems: mean-shift and linear-trend-shift.
Let us consider a random sequence $\bm{X}=(X_1, \ldots, X_n)^\top \sim \mathcal{N}(\bm{\mu}, \Sigma)$, where $n$ is the length of sequence, $\bm{\mu}\in \mathbb{R}^n$ is unknown mean vector, and $\Sigma\in\mathbb{R}^{n\times n}$ is covariance matrix which is known or estimable from external data
\footnote{It is important to note that we do NOT assume the signals are normally distributed, but assume a situation where the noise follows or can be approximated by a normal distribution.}.
Although we do not assume any true structures in the mean vector $\bm{\mu}$, we consider the case where data analysts believe that the sequence can be reasonably approximated by either of mean-shift structure:
\begin{equation}
    \mu_i = c_k\quad (\tau_{k-1}\leq i < \tau_k, k\in[K+1]),
\end{equation}
or linear-trend-shift structure:
\begin{equation}
    \mu_i = t_k i + v_k\quad (\tau_{k-1}\leq i < \tau_k, k\in [K+1]),
\end{equation}
where $K$ is the known number of CPs and $\tau_1, \ldots, \tau_K$ are the true CP locations.
We assume that, in mean-shift problem, $c_k\neq c_{k+1}$ hold for any $k\in[K]$, while, in linear-trend-shift problem, $t_k\neq t_{k+1}$ hold for any $k\in[K]$.
Given an observed sequence $\bm{x}=(x_1, \ldots, x_n)^\top$, the goal of CP detection is to estimate the true CP locations.
%
% The vector of detected CP locations are denoted as $\bm{\tau}^\mathrm{det}=(\tau_1^\mathrm{det}, \ldots,\tau_K^\mathrm{det})$.
%
% We set $\tau_0^\mathrm{det}=0, \tau_{K+1}^\mathrm{det}=n$ and define \red{$\bm{X}_{s:b}$ as a subsequence of $\bm{X}$} from positions $s$ to $b$, where \red{$1\leq s\leq b\leq n$}.
%
\subsection{CP Detection by RNN}
We perform CP detection using a trained-RNN for time-series prediction.
We consider an RNN that can predict the value of the next point from a preceding subsequence of length $l$.
We assume that the RNN has been already trained with a set of time-series data without any CPs.
To detect CPs, we compare the difference between the observed and the predicted subsequence with length $m$, where the long-term prediction is made by the trained-RNN.
Specifically, let $X_{i+1}^\mathrm{pred}$ be the prediction obtained by inputting $\bm{X}_{i-l+1:i}$ into the RNN, where $\bm{X}_{s:b}$ is a subsequence of $\bm{X}$ from positions $s$ to $b$, where $1\leq s\leq b\leq n$.
Then, the prediction $X_{i+2}^\mathrm{pred}$ is similarly obtained by inputting $(\bm{X}_{i-l+2:i}, X_{i+1}^\mathrm{pred})$ into the RNN.
By repeating these operations $m$ times, we similarly obtain $X_{i+1}^\mathrm{pred},\ldots,X_{i+m}^\mathrm{pred}$.

\begin{figure}[htbp]
    \centering
    \includegraphics[width=\linewidth]{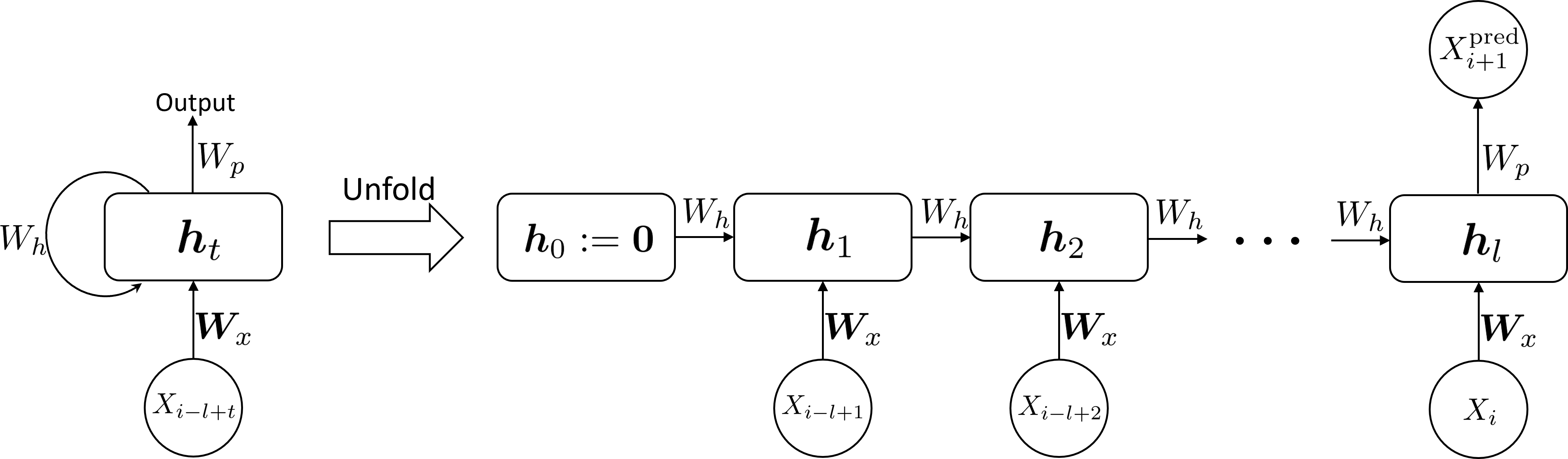}
    \caption{Architecture of RNN considered in this paper}
    \label{fig:rnn}
\end{figure}

We denote the vector of error scores $\bm{e} = (e_1, \ldots, e_n)^\top \in \mathbb{R}^n$ where $e_i, i \in [n]$, is the difference at the $i$-th point.
Using the above notations, the error score is computed as
\footnote{For $i\notin [l\colon n-m]$, the error score is not properly computed and we set $e_i=0$ for convenience.}
\begin{equation}
    e_i = \frac{1}{m}
    \|
    (X_{i+1}^\mathrm{pred},\ldots,X_{i+m}^\mathrm{pred})^\top -
    (X_{i+1},\ldots,X_{i+m})^\top
    \|_2^2,
    i\in[l:n-m].
\end{equation}
%
%
% that predicts value at the next point from a subsequence of length $l$.
%
% First, the predictions made by the RNN are input again to make long-term predictions (we denote the length as $m$) sequentially.
%
% Then, by comparing the results with the actual sequence, we obtain the error scores $\bm{e}=(e_1,\ldots,e_n)^\top\in\mathbb{R}^n$ at each point.
%
Then, we define the anomaly scores $\bm{s}^\mathrm{ano}=(s_1^\mathrm{ano},\ldots,s_n^\mathrm{ano})^\top\in\mathbb{R}^n$ for each point. % by smoothing the error scores $\bm{e}$.
Specifically, the anomaly scores $s_i^\mathrm{ano}, i\in[n]$, are defined by smoothing the error scores $\bm{e}$ with the window size $w$ as
\begin{equation}
    s_i^\mathrm{ano} = \frac{1}{w}\sum_{j=-(w-1)/2}^{(w-1)/2}e_{i+j},
\end{equation}
where $w$ is an odd number.
Based on the anomaly scores $\bm{s}^\mathrm{ano}$, we conduct CP detection as follows.
Let the set of the indices of local maxima of anomaly scores $\bm{s}^\mathrm{ano} := (s^\mathrm{ano}_1, \ldots, s^\mathrm{ano}_n)^\top$, i.e., points at which the anomaly score $s^\mathrm{ano}_i$ is greater than both of the previous one and the next one as
% The set of the indices of local maxima of anomaly scores $\bm{s}^\mathrm{ano}$, $\mathcal{M}$, can be expressed as
%
\begin{equation}
    \mathcal{M} :=
    \left\{
    i\mid
    \sign(s_{i+1}^\mathrm{ano}-s_i^\mathrm{ano})=-1,
    \sign(s_i^\mathrm{ano}-s_{i-1}^\mathrm{ano})=1, i\in[2:n-1]
    \right\}.
\end{equation}
Furthermore, let us define $\mathcal{M}_1=\mathcal{M}$ and $\mathcal{M}_{k}=\mathcal{M}_{k-1}\setminus{\{\argmax_{i\in\mathcal{M}_{k-1}} s_i^\mathrm{ano}\}}$ for $k\geq 2$.
We select $K$ CPs in descending order of local maxima of anomaly score $s^\mathrm{ano}_i$ and represent them as a concatenated $K$-dimensional vector denoted as $\bm{\tau}^\mathrm{det}=(\tau_1^\mathrm{det},\ldots,\tau_K^\mathrm{det})\in\{1,n\}^K$ with $\tau_0^\mathrm{det}=0$ and $\tau_{K+1}^\mathrm{det}$ for notational consistency.
Namely, using the notations $\mathcal{M}_1, \ldots, \mathcal{M}_K$, the elements of the vector $\bm{\tau}^\mathrm{det}$ is defined as $\left\{\argmax_{i\in\mathcal{M}_k}s_i^\mathrm{ano} \mid k\in[K]\right\}$ in descending order.
% $\bm{\tau}^\mathrm{det}$ is given as a vector of elements of $\left\{\argmax_{i\in\mathcal{M}_k}s_i^\mathrm{ano} \mid k\in[K]\right\}$ in descending order.
%
Schematic illustrations of the problem setups considered in this paper are provided in Figure~\ref{fig:schematic} (Figure~\ref{fig:schematic_ms} for mean-shift problem , Figure~\ref{fig:schematic_lt} for linear-trend-shift problem).
%
% We denote the operation of obtaining T by applying these procedures to a series X as the mapping B:X\to T .
We denote the event that the CP locations vector $\bm{\tau}^\mathrm{det}$ is detected by these algorithm $\mathcal{A}$ from the sequence $\bm{X}$ as
\begin{equation}
    \mathcal{A}:\bm{X} \mapsto \bm{\tau}^\mathrm{det}.
\end{equation}
\paragraph{Time-Series Predictions by RNN}
Finally, we formulate the computational process to obtain the prediction $X_{i+1}^\mathrm{pred}\in\mathbb{R}$ by inputting $\bm{X}_{i-l+1:i}\in\mathbb{R}^l$ to the RNN.
We consider a standard RNN as shown in Figure~\ref{fig:rnn} in this paper.
The RNN has a $d^h$-dimensional internal state $\bm{h}_{t-1}\in\mathbb{R}^n$, which is updated to $\bm{h}_{t}\in\mathbb{R}^n$ by combining it with the one element of input $X_{i-l+t}\in\mathbb{R}$, for each time step $t\in[l]$.
Let the initial value of the internal state $\bm{h}_0$ be $\bm{0}$.
As the prediction $X_{i+1}^\mathrm{pred}$, we use the output obtained by affine transformation of the final internal state $\bm{h}_l$.
To summarize, using the activation function $f$, the weights $W_h\in\mathbb{R}^{d^h\times d^h}$ for the internal state, the weights $\bm{W}_x\in\mathbb{R}^{d^h}$ for the each input, the bias term $\bm{W}_b\in\mathbb{R}^{d^h}$, and the weight $W^p\in\mathbb{R}^{1\times d^h}$ to obtain the prediction, the process of RNN can be formulated as
\begin{gather}
    \bm{h}_{0} = \bm{0},\\
    \bm{h}_t = f(W_h\bm{h}_{t-1}+\bm{W}_xX_{i-l+t} + \bm{W}_b), \forall t\in[l], \\
    X_{i+1}^\mathrm{pred} = W_p\bm{h}_l.
\end{gather}
\subsection{Statistical Inference on CPs}
\begin{figure}[htbp]
    \begin{minipage}[b]{\linewidth}
        \centering
        \includegraphics[width=\linewidth]{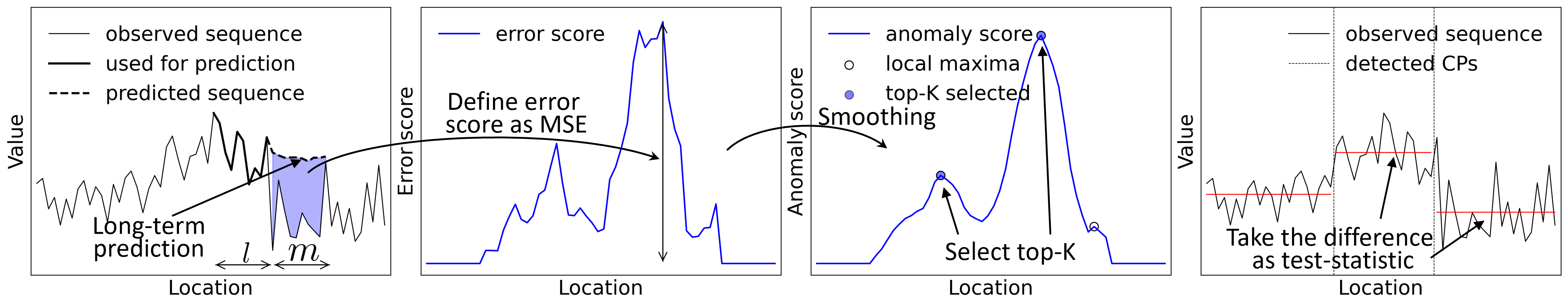}
        \subcaption{mean-shift}
        \label{fig:schematic_ms}
    \end{minipage}
    \begin{minipage}[b]{\linewidth}
        \centering
        \includegraphics[width=\linewidth]{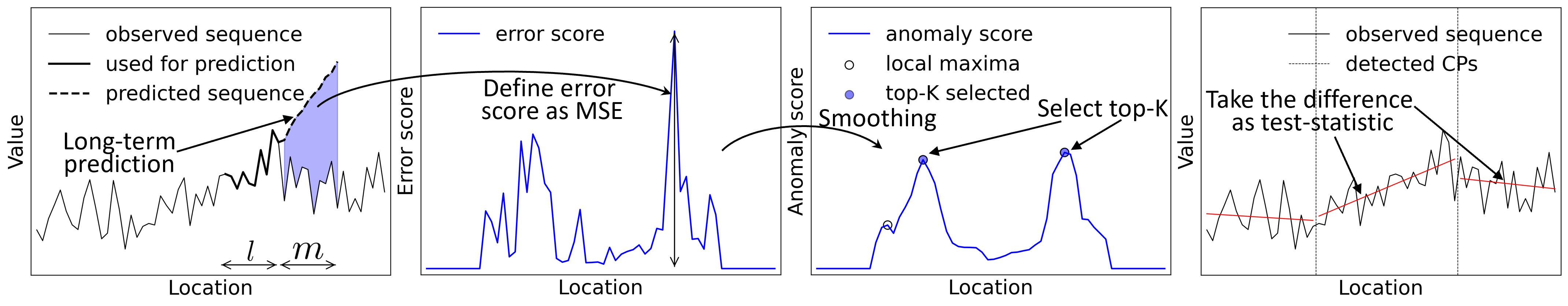}
        \subcaption{linear-trend-shift}
        \label{fig:schematic_lt}
    \end{minipage}
    \caption{
        Schematic illustration of the two change-point detection problems (mean-shift and linear-trend-shift problems) considered in this study.
        We define the error score for each point by the long-term prediction error by RNN.
        By smoothing the error scores, the anomaly scores are defined, and the CPs are detected by selecting the maxima with top-$K$.
        We evaluate the statistical significance of the detected CPs by testing the difference between the mean/linear-trend before and after the each detected CP.
    }
    \label{fig:schematic}
\end{figure}

To quantify the statistical significance of the $k$-th detected CP $\tau_k^\mathrm{det}$, we consider the two types of statistical tests for mean-shift and linear-trend-shift, respectively.
The first type, which we call, mean-shift problem is based on the belief that the mean is constant in the each segment before and after the detected CP.
That is, for $c(\tau_{k-1}^\mathrm{det}:\tau_{k}^\mathrm{det}):=\mu_{\tau_{k-1}^\mathrm{det}+1}=\cdots=\mu_{\tau_k^\mathrm{det}}$ and $c(\tau_{k}^\mathrm{det}:\tau_{k+1}^\mathrm{det}):=\mu_{\tau_{k}^\mathrm{det}+1}=\cdots=\mu_{\tau_{k+1}^\mathrm{det}}$, we consider the statistical test with the following null and alternative hypotheses:
\begin{equation}
    \mathrm{H}_{0, k}:
    c(\tau_{k-1}^\mathrm{det}:\tau_{k}^\mathrm{det})
    =
    c(\tau_{k}^\mathrm{det}:\tau_{k+1}^\mathrm{det}),
    \
    \text{v.s.},
    \
    \mathrm{H}_{1, k}:
    c(\tau_{k-1}^\mathrm{det}:\tau_{k}^\mathrm{det})
    \neq
    c(\tau_{k}^\mathrm{det}:\tau_{k+1}^\mathrm{det}).
\end{equation}
The second type, which we call, linear-trend-shift problem is based on the belief that the value change linearly with a constant slope in the each segment before and after the detected CP.
That is, for $t(\tau_{k-1}^\mathrm{det}:\tau_{k}^\mathrm{det}):=\mu_{\tau_{k-1}^\mathrm{det}+2}-\mu_{\tau_{k-1}^\mathrm{det}+1}=\cdots=\mu_{\tau_k^\mathrm{det}}-\mu_{\tau_k^\mathrm{det}-1}$ and $t(\tau_{k-1}^\mathrm{det}:\tau_{k}^\mathrm{det}):=\mu_{\tau_{k}^\mathrm{det}+2}-\mu_{\tau_{k}^\mathrm{det}+1}=\cdots=\mu_{\tau_{k+1}^\mathrm{det}}-\mu_{\tau_{k+1}^\mathrm{det}-1}$, we consider the statistical test with the following null and alternative hypotheses:
\begin{equation}
    \mathrm{H}_{0, k}:
    t(\tau_{k-1}^\mathrm{det}:\tau_{k}^\mathrm{det})
    =
    t(\tau_{k}^\mathrm{det}:\tau_{k+1}^\mathrm{det}),
    \
    \text{v.s.},
    \
    \mathrm{H}_{1, k}:
    t(\tau_{k-1}^\mathrm{det}:\tau_{k}^\mathrm{det})
    \neq
    t(\tau_{k}^\mathrm{det}:\tau_{k+1}^\mathrm{det}).
\end{equation}
\paragraph{Test-statistic}
In the case of mean-shift problem, it is reasonable to define the test-statistic by taking the difference of the average of the two segments before and after the $k$-th detected CP, i.e., we define the test-statistic as
\begin{equation}
    \label{eq:test-statistic_mean-shift}
    \bm{\eta}_k^\top \bm{X} =
    \frac{1}{\tau_{k}^\mathrm{det}-\tau_{k-1}^\mathrm{det}}
    \sum_{j=\tau_{k-1}^\mathrm{det}+1}^{\tau_{k}^\mathrm{det}}X_j
    -
    \frac{1}{\tau_{k+1}^\mathrm{det}-\tau_{k}^\mathrm{det}}
    \sum_{j=\tau_{k}^\mathrm{det}+1}^{\tau_{k+1}^\mathrm{det}}X_j,
\end{equation}
where $\bm{\eta}_k=\frac{1}{\tau_{k}^\mathrm{det}-\tau_{k-1}^\mathrm{det}}\bm{1}_{\tau_{k-1}^\mathrm{det}+1:\tau_{k}^\mathrm{det}}^n-\frac{1}{\tau_{k+1}^\mathrm{det}-\tau_{k}^\mathrm{det}}\bm{1}_{\tau_{k}^\mathrm{det}+1:\tau_{k+1}^\mathrm{det}}^n$ and $\bm{1}_{s:e}^n\in\mathbb{R}^n$ is a vector whose elements from position $s$ to $e$ are set to $1$ and $0$ otherwise (see Figure~\ref{fig:schematic_ms}).
In the case of linear-trend-shift problem, it is reasonable to define the test-statistic by taking the difference of the regression coefficients from the least squares method of the two segments before and after the $k$-th detected CP, i.e., we define the test-statistic as
\begin{equation}
    \label{eq:test-statistic_linear-trend-shift}
    \bm{\eta}_k^\top \bm{X}
    =
    \bm{g}_{\tau_{k-1}^\mathrm{det}+1:\tau_{k-1}^\mathrm{det}}^\top
    \bm{X}_{\tau_{k-1}^\mathrm{det}+1:\tau_{k}^\mathrm{det}}
    -
    \bm{g}_{\tau_{k}^\mathrm{det}+1:\tau_{k+1}^\mathrm{det}}^\top
    \bm{X}_{\tau_{k}^\mathrm{det}+1:\tau_{k+1}^\mathrm{det}},
\end{equation}
where $\bm{\eta}_k=\bm{g}_{\tau_{k-1}^\mathrm{det}+1:\tau_{k}^\mathrm{det}}^n-\bm{g}_{\tau_{k}^\mathrm{det}+1:\tau_{k}^\mathrm{det}}^n$ and $\bm{g}_{s:b}^n\in\mathbb{R}^n$ is a vector whose elements from position $s$ to $b$ are embedded vector $\bm{g}_{s:b}$ and set to $0$ otherwise, and $\bm{g}_{s:b}\in\mathbb{R}^{b-s+1}$ is a vector defined as $(\bm{g}_{s:b})_i=6(2i-b+s-2)/\{(b-s)(b-s+1)(b-s+2)\}$ (see Figure~\ref{fig:schematic_lt}).
\paragraph{Conditional Selective Inference.}
To conduct the above statistical tests, we consider the test-statistic conditional on the event that the same detected CPs are obtained, i.e.,
\begin{equation}
    \label{eq:conditional_distribution}
    \bm{\eta}_k^\top \bm{X} \mid \{\mathcal{A}(\bm{X}) = \mathcal{A}(\bm{x})\}.
\end{equation}
To compute the selective $p$-value based on the conditonal sampling distribution in \eqref{eq:conditional_distribution}, we need to additionally condition on the nuisance component $\mathcal{Q}(\bm{X})$ defined as
\begin{equation}
    \mathcal{Q}(\bm{X}) =
    \left(
    I_n - \frac{\Sigma\bm{\eta}_k\bm{\eta}_k^\top}{\bm{\eta}_k^\top\Sigma\bm{\eta}_k}
    \right)
    \bm{X}.
\end{equation}
The selective $p$-value is then computed as
\begin{equation}
    \label{eq:selective_p-value}
    p^\mathrm{selective}_k = \mathbb{P}_{\mathrm{H}_{0,k}}
    \left(
    \left|\bm{\eta}_k^\top \bm{X}\right| \geq
    \left|\bm{\eta}_k^\top \bm{x}\right| \mid
    \bm{X} \in \mathcal{X}
    \right)
\end{equation}
where $\mathcal{X}=\{\bm{X}\in\mathbb{R}^n\mid \mathcal{A}(\bm{X})=\mathcal{A}(\bm{x}), \mathcal{Q}(\bm{X})=\mathcal{Q}(\bm{x})\}$.
In \eqref{eq:selective_p-value}, $\mathcal{A}(\bm{X})=\mathcal{A}(\bm{x})$ indicates the event that the detected CP locations vector for a random sequence $\bm{X}$ is the same as the detected CP locations vector for the observed sequence $\bm{x}$.
\paragraph{Characterization of the Conditional Data Space.}
The data space $\mathbb{R}^n$ conditional on $\mathcal{A}(\bm{X})=\mathcal{A}(\bm{x})$ is represented as a subset of $\mathbb{R}^n$.
Furthermore, by conditioning also on the nuisance component $\mathcal{Q}(\bm{X})=\mathcal{Q}(\bm{x})$, the subspace $\mathcal{X}$ in \eqref{eq:selective_p-value} is a one-dimensional line in $\mathbb{R}^n$.
Therefore, the set $\mathcal{X}$ can be re-written, using a scalar parameter $z\in\mathbb{R}$, as
\begin{equation}
    \mathcal{X} = \{\bm{X}(z)\in\mathbb{R}^n\mid \bm{X}(z)=\bm{a}+\bm{b}z, z\in\mathcal{Z}\}
\end{equation}
where vectors $\bm{a}$ and $\bm{b}$ are defined as
\begin{equation}
    \label{eq:direction_vectors}
    \bm{a} = \mathcal{Q}(\bm{x}),
    \bm{b} = \frac{\Sigma\bm{\eta}_k\bm{\eta}_k^\top}{\bm{\eta}_k^\top\Sigma\bm{\eta}_k}\bm{x},
\end{equation}
and the region $\mathcal{Z}$ is defined as
\begin{equation}
    \label{eq:target_region}
    \mathcal{Z} = \{z\in\mathbb{R}\mid \mathcal{A}(\bm{a}+\bm{b}z) = \mathcal{A}(\bm{x})\}.
\end{equation}
Let us consider a random variable $Z\in\mathbb{R}$ and its observation $Z^\mathrm{obs}\in\mathbb{R}$ such that they respectively satisfy $\bm{X}=\bm{a}+\bm{b}Z$ and $\bm{x}=\bm{a}+\bm{b}Z^\mathrm{obs}$.
The selective $p$-value in \eqref{eq:selective_p-value} is re-written as
\begin{equation}
    \label{eq:selective_p_from_region}
    p^\mathrm{selective}_k = \mathbb{P}_{\mathrm{H}_{0,k}}
    \left(\left|Z\right| \geq \left|Z^\mathrm{obs}\right|\mid Z\in\mathcal{Z}\right).
\end{equation}
Because the unconditional variable $Z\sim \mathcal{N}(0, \bm{\eta}_k^\top\Sigma\bm{\eta}_k)$ under the null hypothesis, the conditional random variable $Z\mid Z\in\mathcal{Z}$ follows a truncated normal distribution.
Once the truncation region $\mathcal{Z}$ is identified, the selective $p$-value in \eqref{eq:selective_p_from_region} can be easily computed.
Thus, the remaining task is reduced to the characterization of $\mathcal{Z}$.

\newpage
\section{Proposed Method}
\label{sec:sec3}
In general, conditioning event only on the result of algorithm i.e., $\mathcal{A}(\bm{a}+\bm{b}z)=\mathcal{A}(\bm{x})$ is too complicated to characterize due to the complexity of the trained-RNN.
Therefore, in SI literature, it is common to introduce additional conditioning on events in the process of algorithm $\mathcal{A}$ for tractability.
This additional conditioning is often reffered to as \emph{over-conditioning} because they are redundant for valid inference, but are necessary for computational tractability.
Note that SIs with over-conditioning are still valid in the sense that the type I error rate can be controlled at the desired significance level.
However, over-conditioning tend to decrease the power of SI.
In this study, we first consider SI with over-conditioning, and then remove all the redundant conditions afterward.
\subsection{Characterizing Selection Event with Over-Conditioning}
\begin{figure}[htbp]
    \centering
    \includegraphics[width=0.6\linewidth]{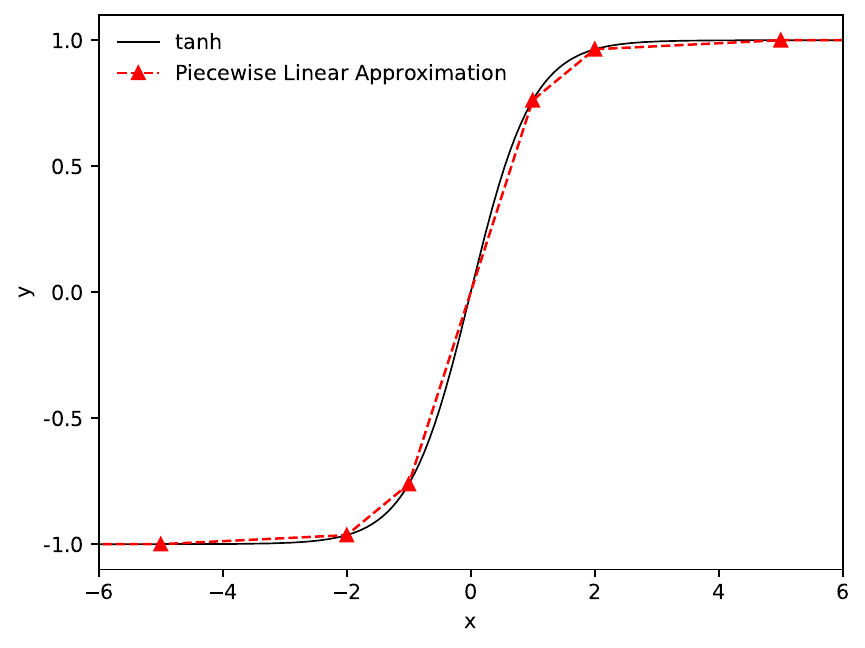}
    \caption{Piecewise linear approximation of hyperbolic tangent function}
    \label{fig:tanh}
\end{figure}

Since we only need to consider a line in the data space in $\mathcal{Z}$, with a slight abuse of notations, we interpret that a real number $r$ is an input to algorithm $\mathcal{A}$, i.e., we write $\mathcal{A}(r)=\mathcal{A}(\bm{a}+\bm{b}r)$.
Specifically, we consider the over-conditioned region defined as
\begin{gather}
    \label{eq:oc_region}
    \mathcal{Z}^\mathrm{oc}(\bm{a}+\bm{b}z)\\
    =\{
    r\in\mathbb{R}\mid
    \mathcal{S}_\mathrm{active}(r)=\mathcal{S}_\mathrm{active}(z),
    \mathcal{S}_\mathrm{sign}(r)=\mathcal{S}_\mathrm{sign}(z),
    \mathcal{S}_\mathrm{sort}(r)=\mathcal{S}_\mathrm{sort}(z)
    \},
\end{gather}
where $\mathcal{S}_\mathrm{active}$, $\mathcal{S}_\mathrm{sign}$, and $\mathcal{S}_\mathrm{sort}$ are events characterized in the process of algorithm $\mathcal{A}$ (i.e., the computation process for detecting CPs) whose details will be explained later.
These events are over-conditioned event because it is a proper subset of the region corresponding to the minimum conditioning, i.e.,
\begin{equation}
    \mathcal{Z}^\mathrm{oc}(\bm{a}+\bm{b}z)
    \subsetneq
    \{r\in\mathbb{R}\mid \mathcal{A}(r)=\mathcal{A}(z)\}.
\end{equation}
%
% Note that these conditionals include the condition $\mathcal{A}(z)=\mathcal{A}(r)$.
%
\paragraph{Over-Conditioning on ``active''-ness of activation functions}
RNN is a network for nonlinear transformation that recursively applies affine operations and activation functions.
However, if a piecewise-linear function such as ReLU or Leaky ReLU is used as the activation function, the entire process can be represented as a piecewise-linear function.
This means that, if we condition on the activeness of each activation function (e.g., in the case of ReLU, whether the input to the activation function is at the constant part or the linear part), the computation of the trained-RNN is represented as a linear operation with respect to its input.
When non-piecewise-linear activation functions, such as hyperbolic tangent functions, are used as activation functions in RNN, the same discussion can be made by considering piecewise linear approximation (see Figure~\ref{fig:tanh}).

To discuss the region $\{r\in\mathbb{R}\mid\mathcal{S}_\mathrm{active}(r)=\mathcal{S}_\mathrm{active}(z)\}$, we consider a sequence parametrized as $\bm{X}(r)=\bm{a}+\bm{b}r$.
For any $i\in[l:n-m]$, assuming for any $t\in[l]$ that internal state $\bm{h}_{t-1}(r)$ is expressed as a linear expression of $r$ with $\bm{h}_{t-1}(r)=\bm{a}_{t-1}^f+\bm{b}_{t-1}^fr$, the following transformation of the equation
\begin{align}
    \bm{h}_t(r) & = f\left(W_h(\bm{a}_{t-1}^f+\bm{b}_{t-1}^fr)+\bm{W}_x(a_{i-l+t}+b_{i-l+t}r) + \bm{W}_b\right)          \\
                & = f\left((W_h\bm{a}_{t-1}^f\bm{W}_xa_{i-l+t} + \bm{W}_b)+(W_h\bm{b}_{t-1}^f+\bm{W}_xb_{i-l+t})r\right) \\
                & = f(\bm{a}_t^h + \bm{b}_t^hr)
\end{align}
holds, where $\bm{a}_t^h=W_h\bm{a}_{t-1}^f\bm{W}_xa_{i-l+t} + \bm{W}_b$ and $\bm{b}_t^h=W_h\bm{b}_{t-1}^f+\bm{W}_xb_{i-l+t}$.
Thus, conditional on all elements of $\bm{a}_1^h+\bm{b}_1^hr$ and $\bm{a}_1^h+\bm{b}_1^hz$ being equally active or inactive, the piecewise-linear activation function is an affine operation and is expressed as
\begin{equation}
    \bm{h}_t(r) = f(\bm{a}_t^h + \bm{b}_t^hz) = (W_f\bm{a}_t^h) + (W_f\bm{b}_t^h)r.
\end{equation}
Therefore, again updated internal state $\bm{h}_t(r)$ is also a linear expression of $r$.
Also, when $t=1$, $\bm{h}_0=\bm{0}+\bm{0}z$, which satisfies the assumption, so if the above process is applied recursively, the prediction is also a linear expression of $r$, and becomes a linear expression of $r$ as
\begin{equation}
    X_{i+1}^\mathrm{pred}(r) = W_p\bm{h}_l = (W_pW_f\bm{a}_l^h) + (W_pW_f\bm{b}_l^h)r.
\end{equation}
Thus, these operations can be repeated for each long-term prediction, and if applied $m$ times sequentially, we obtain
\begin{equation}
    (X_{i+1}^\mathrm{pred}(r),\ldots,X_{i+m}^\mathrm{pred}(r))^\top
    = \bm{a}^\mathrm{pred} + \bm{b}^\mathrm{pred}r.
\end{equation}
as a linear expression of $r$, where $\bm{a}^\mathrm{pred}$ and $\bm{b}^\mathrm{pred}$ are composed of $W_pW_f\bm{a}_l^h$ and $W_pW_f\bm{b}_l^h$ for each sequential process side by side.
Using these notations, the error score $e_i(r)$ can be expressed as a quadratic expression of $r$ such that
\begin{align}
    e_i(r) & =
    \frac{1}{m}
    \|
    (\bm{a}^\mathrm{pred} + \bm{b}^\mathrm{pred}r) -
    (\bm{a}_{i+1:i+m}+\bm{b}_{i+1:i+m}r)
    \|_2^2                                         \\
           & = \alpha_i r^2 + \beta_i r +\gamma_i,
\end{align}
where $\alpha_i=(\bm{b}^\mathrm{pred}-\bm{b}_{i+1:i+m})^\top(\bm{b}^\mathrm{pred}-\bm{b}_{i+1:i+m})/m$, $\beta_i=2(\bm{b}^\mathrm{pred}-\bm{b}_{i+1:i+m})^\top(\bm{a}^\mathrm{pred}-\bm{a}_{i+1:i+m})/m$, and $\gamma_i= (\bm{a}^\mathrm{pred}-\bm{a}_{i+1:i+m})^\top(\bm{a}^\mathrm{pred}-\bm{a}_{i+1:i+m})/m$.

To summarize the above discussion, we define the region $\{r\in\mathbb{R}\mid\mathcal{S}_\mathrm{active}(r)=\mathcal{S}_\mathrm{active}(z)\}$ as
\begin{gather}
    \{r\in\mathbb{R}\mid\mathcal{S}_\mathrm{active}(r)=\mathcal{S}_\mathrm{active}(z)\}\\
    \label{eq:oc_on_activate}
    = \bigcap_{i=l}^{n-m}\bigcap_{j=1}^m\bigcap_{t=1}^l
    \{r\in\mathbb{R}\mid \bm{s}_j(\bm{a}_t^h+\bm{b}_t^hr)=\bm{s}_j(\bm{a}_t^h+\bm{b}_t^hz)\}
\end{gather}
where $\bm{s}_j(\bm{a}_t^h+\bm{b}_t^hr)$ is a vector that represents whether each is active or inactive for all elements of the $t$-th internal state at the $j$-th long-term prediction.
For example, if the activation function is ReLU, then $\bm{s}_j$ is just a vectorization of the sign function.
Each set in \eqref{eq:oc_on_activate} is generally expressed as a linear inequality.
For any $r$ which satisfy $\mathcal{S}_\mathrm{active}(r)=\mathcal{S}_\mathrm{active}(z)$,
\begin{equation}
    \label{eq:quadratic_error_score}
    \bm{e}(r)= \bm{\alpha}r^2 + \bm{\beta}r + \bm{\gamma}
\end{equation}
holds by setting $\alpha_i=\beta_i=\gamma_i=0$ for $i\notin [l:n-m]$.
\paragraph{Over-Conditioning on ``signs'' of anomaly score changes}
When conditioning on $\mathcal{S}_\mathrm{active}$, the error scores are expressed in a quadratic expression of $r$ as in \eqref{eq:quadratic_error_score}.
Therefore, the anomaly scores are also clearly quadratic in $r$.
Namely, it can be expressed as
\begin{equation}
    \label{eq:quadratic_anomaly_score}
    \bm{s}^\mathrm{ano}(r) = \bm{\alpha}r^2 + \bm{\beta}r + \bm{\gamma}
\end{equation}
with a slight abuse of notations.
Next, we want to condition on the set of the indices of local maxima of the anomaly scores, i.e., $\mathcal{M}(r)=\mathcal{M}(z)$.
For this purpose, it is sufficient to condition on the signs of all the differences between adjacent anomaly scores, so we can define the region $\{r\in\mathbb{R}\mid\mathcal{S}_\mathrm{sign}(r)=\mathcal{S}_\mathrm{sign}(z)\}$ as
\begin{gather}
    \{r\in\mathbb{R}\mid
    \mathcal{S}_\mathrm{sign}(r)=\mathcal{S}_\mathrm{sign}(z)\}\\
    \label{eq:oc_on_sign}
    = \bigcap_{i=l}^{n-m-1}
    \{
    r \in \mathbb{R}\mid
    \sign(s_{i+1}^\mathrm{ano}(r)-s_i^\mathrm{ano}(r))=
    \sign(s_{i+1}^\mathrm{ano}(z)-s_i^\mathrm{ano}(z))
    \}.
\end{gather}
If we condition on $\mathcal{S}_\mathrm{active}$, from \eqref{eq:quadratic_anomaly_score}, each set in \eqref{eq:oc_on_sign} can be expressed as a quadratic inequality.
\paragraph{Over-Conditioning on ``sort''}
Finally, we have to condition on the operation of selecting the top-$K$ elements from the set of local maxima.
To this end, it is sufficient to condition on a large/small relationship for all elements of the set of local maxima.
If we order $\mathcal{M}(z) = \{m(1), \ldots, m(|\mathcal{M}|)\}$, the set of local maxima is $\{s_{m(i)}^\mathrm{ano}\mid i\in[|\mathcal{M}(z)|]\}$.
We define the permutation $\sigma^z$ as $s_{m(i)}^\mathrm{ano}$ being the $\sigma^z(i)$-th largest in this set.
Using this permutation, we can define the region $\{r\in\mathbb{R}\mid\mathcal{S}_\mathrm{sort}(r)=\mathcal{S}_\mathrm{sort}(z)\}$ as
\begin{gather}
    \{r\in\mathbb{R}\mid
    \mathcal{S}_\mathrm{sort}(r)=\mathcal{S}_\mathrm{sort}(z)\}\\
    \label{eq:oc_on_sort}
    = \bigcap_{i=1}^{|\mathcal{M}|-1}
    \{
    r \in \mathbb{R}\mid
    s_{m(\sigma^z(i+1))}^\mathrm{ano} \geq s_{m(\sigma^z(i))}^\mathrm{ano}
    \}.
\end{gather}
If we condition on $\mathcal{S}_\mathrm{active}$, from \eqref{eq:quadratic_anomaly_score}, each set in \eqref{eq:oc_on_sign} can be expressed as a quadratic inequality.
Furthermore, by conditioning on $\mathcal{S}_\mathrm{sign}$, each set in \eqref{eq:oc_on_sort} appropriately represent the large/small relationship of maxima.
\subsection{Characterizing Selection Event for Minimum-Conditioning via Parametric Programming}
Since $\mathcal{Z}^\mathrm{oc}(\bm{a}+\bm{b}z)$ in \eqref{eq:oc_region} is well defined by the previous discussion, it is sufficient to consider the union of them to obtain $\mathcal{Z}$ in \eqref{eq:target_region} as follows
\begin{equation}
    \label{eq:obtaine_region_by_union}
    \mathcal{Z} = \bigcup_{z\in\mathbb{R}\mid \mathcal{A}(z)=\mathcal{A}(\bm{x})}
    \mathcal{Z}^\mathrm{oc}(\bm{a}+\bm{b}z).
\end{equation}
From the above discussion, $\mathcal{Z}^\mathrm{oc}(\bm{a}+\bm{b}z)$ is given by the intersection of a finite number of linear and quadratic inequalities.
Thus, solving them analytically, it can be represented as a union of multiple non-overlapping intervals, which we denote by
\begin{equation}
    \mathcal{Z}^\mathrm{oc}(\bm{a}+\bm{b}z) =
    \bigcup_{v=1}^{n(\mathcal{S}(z))}[\ell_v^{\mathcal{S}(z)}, u_v^{\mathcal{S}(z)}],
\end{equation}
where $\mathcal{S}(z)=(\mathcal{S}_\mathrm{active}(z), \mathcal{S}_\mathrm{sign}(z), \mathcal{S}_\mathrm{sort}(z))$ and $n(\mathcal{S}(z))$ is the number of intervals determined by $\mathcal{S}(z)$.

The entire SI method for computing selective $p$-values of the detected CPs is summarized in Procedure~\ref{alg:SI}.
Furthermore, Procedure~\ref{alg:compute_path} shows how to efficiently obtain the target region $\mathcal{Z}$ to compute $p_k^\mathrm{selective}$ with a finite number of operations based on \eqref{eq:obtaine_region_by_union}.
Note that while the number of operations required for this procedure is certainly finite, they can be computationally expensive in practise.
Therefore, in practical implementation, the upper and lower bounds of $p_k^\mathrm{selective}$ are obtained from $S_i$ and $R_i$ by using the technique described in \citep{shiraishi2023bounded}, where the while loop is terminated when they become sufficiently tight.
\begin{algorithm}
    \caption{SI for detected CPs}
    \label{alg:SI}
    \begin{algorithmic}[1]
        \REQUIRE $\bm{x}$ and $K$
        \STATE $\bm{\tau}^\mathrm{det}\leftarrow \mathcal{A}(\bm{x})$
        \FOR{$k\in[K]$}
        \STATE $\mathcal{Z}\leftarrow \mathtt{compute\_solution\_path}(\bm{x}, \bm{\tau}^\mathrm{det}, k)$
        \STATE Compute $p_k^\mathrm{selective}$ by \eqref{eq:selective_p_from_region}
        \ENDFOR
        \ENSURE $\{(\tau_k^\mathrm{det}, p_k^\mathrm{selective})\}_{k\in[K]}$
    \end{algorithmic}
\end{algorithm}
\begin{algorithm}
    \caption{\texttt{compute\_solution\_path}}
    \label{alg:compute_path}
    \begin{algorithmic}[1]
        \REQUIRE $\bm{x}$, $\bm{\tau}^\mathrm{det}$ and $k$
        \STATE Compute $\bm{a}$ and $\bm{b}$ by \eqref{eq:direction_vectors}
        \STATE Obtain $Z^\mathrm{obs}$ such that $\bm{x}=\bm{a}+\bm{b}Z^\mathrm{obs}$
        \STATE $S_1 \leftarrow R_1 \leftarrow
            \cup_{v=1}^{n(\mathcal{S}(Z^\mathrm{obs}))}[\ell_v^{\mathcal{S}(Z^\mathrm{obs})}, u_v^{\mathcal{S}(Z^\mathrm{obs})}]$
        \STATE $i\leftarrow 1$
        \WHILE{$S_i^c\neq\emptyset$}
        \STATE Select a $z \in S_i^c$
        % \STATE Compute $\mathcal{A}(z)$ and $\mathcal{S}(z)$
        \STATE $S_{i+1} \leftarrow
            S_i \cup \left\{
            \cup_{v=1}^{n(\mathcal{S}(z))}[\ell_v^{\mathcal{S}(z)}, u_v^{\mathcal{S}(z)}]\right\}$
        \IF{$\mathcal{A}(z) = \mathcal{A}(Z^\mathrm{obs})$}
        \STATE $R_{i+1} \leftarrow
            R_i \cup \left\{
            \cup_{v=1}^{n(\mathcal{S}(z))}
            [\ell_{v}^{\mathcal{S}(z)}, u_{v}^{\mathcal{S}(z)}]
            \right\}$
        \ENDIF
        \STATE $i \leftarrow i+1$
        \ENDWHILE
        \ENSURE $R_{i}(=\mathcal{Z})$
    \end{algorithmic}
\end{algorithm}

\newpage
\section{Experiment}
\label{sec:sec4}
\begin{figure}[htbp]
    % \begin{minipage}[b]{0.49\linewidth}
    \begin{minipage}[b]{0.49\linewidth}
        \centering
        \includegraphics[width=1.0\linewidth]{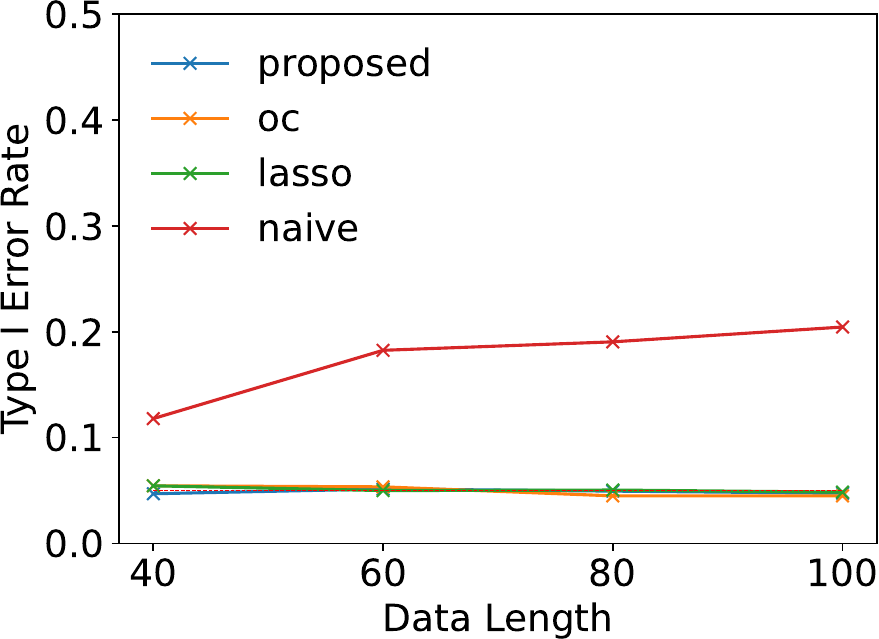}
        \subcaption{Independence}
    \end{minipage}
    \begin{minipage}[b]{0.49\linewidth}
        \centering
        \includegraphics[width=1.0\linewidth]{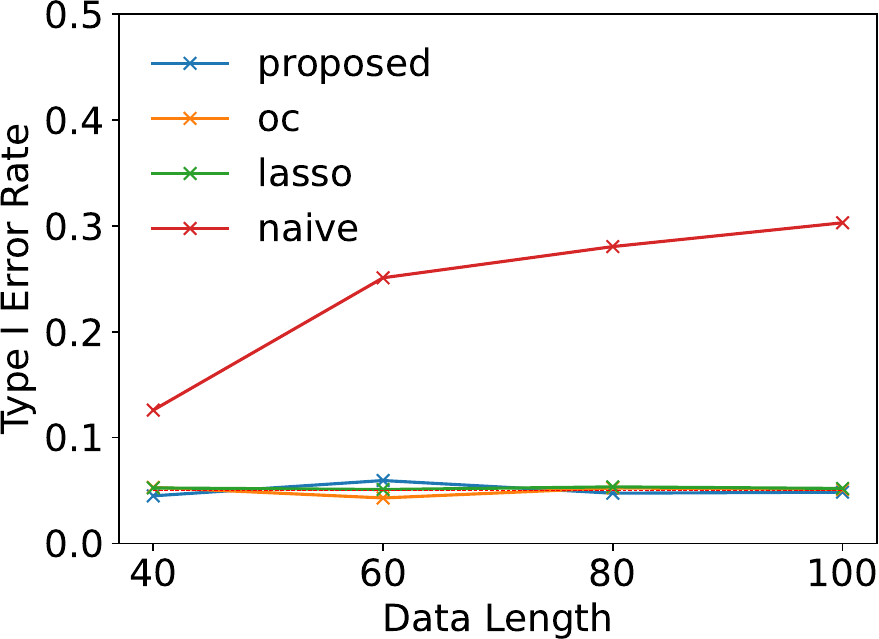}
        \subcaption{Correlation}
    \end{minipage}
    \caption{Type I Error Rate (mean-shift)}
    \label{fig:fpr-mean-shift}
    % \end{minipage}
\end{figure}

\begin{figure}[htbp]
    % \begin{minipage}[b]{0.49\linewidth}
    \begin{minipage}[b]{0.49\linewidth}
        \centering
        \includegraphics[width=1.0\linewidth]{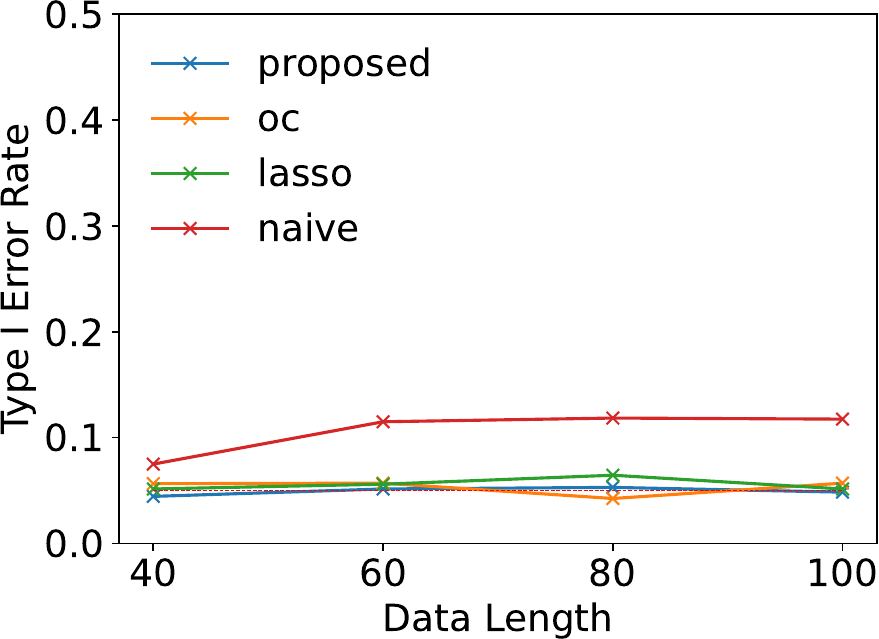}
        \subcaption{Independence}
    \end{minipage}
    \begin{minipage}[b]{0.49\linewidth}
        \centering
        \includegraphics[width=1.0\linewidth]{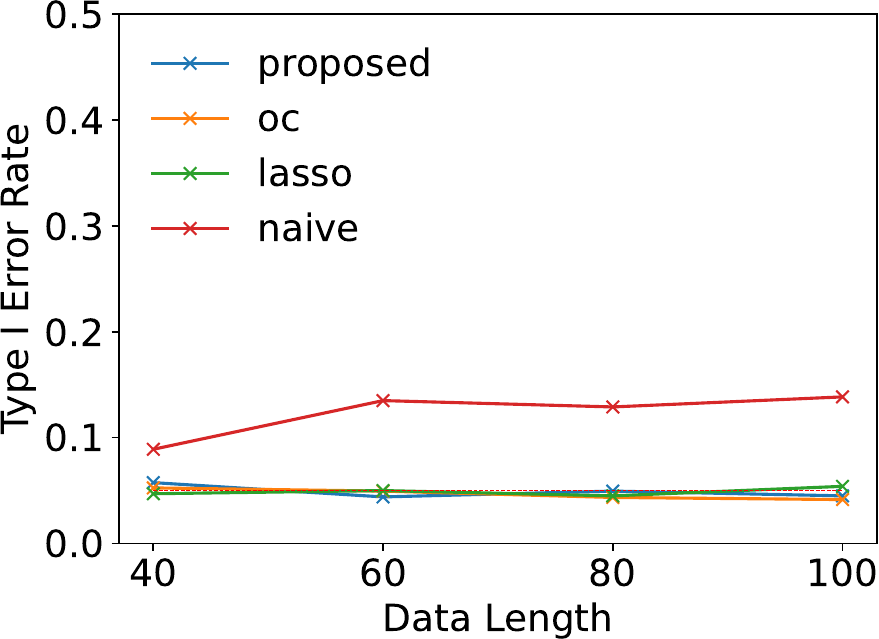}
        \subcaption{Correlation}
    \end{minipage}
    \caption{Type I Error Rate (linear-trend-shift)}
    \label{fig:fpr-linear-trend-shift}
    % \end{minipage}
\end{figure}

\begin{figure}[htbp]
    % \begin{minipage}[b]{0.49\linewidth}
    \begin{minipage}[b]{0.49\linewidth}
        \centering
        \includegraphics[width=1.0\linewidth]{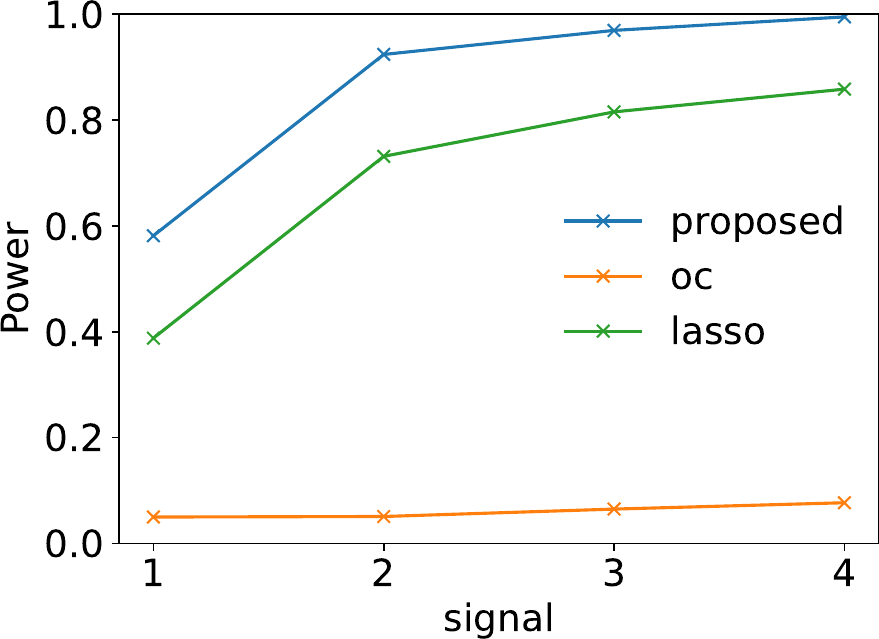}
        \subcaption{Independence}
    \end{minipage}
    \begin{minipage}[b]{0.49\linewidth}
        \centering
        \includegraphics[width=1.0\linewidth]{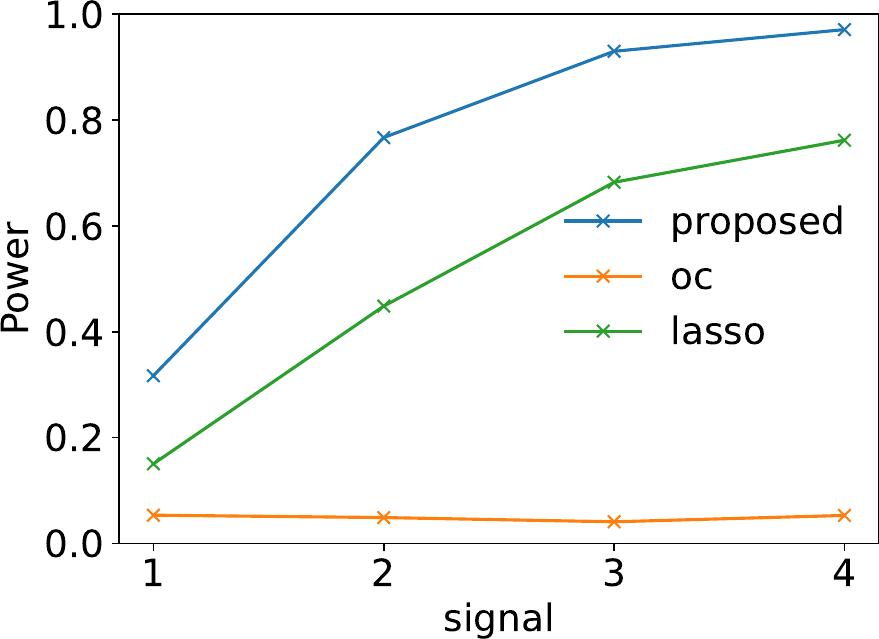}
        \subcaption{Correlation}
    \end{minipage}
    \caption{Power (mean-shift)}
    \label{fig:tpr-mean-shift}
    % \end{minipage}
\end{figure}

\begin{figure}[htbp]
    % \begin{minipage}[b]{0.49\linewidth}
    \begin{minipage}[b]{0.49\linewidth}
        \centering
        \includegraphics[width=1.0\linewidth]{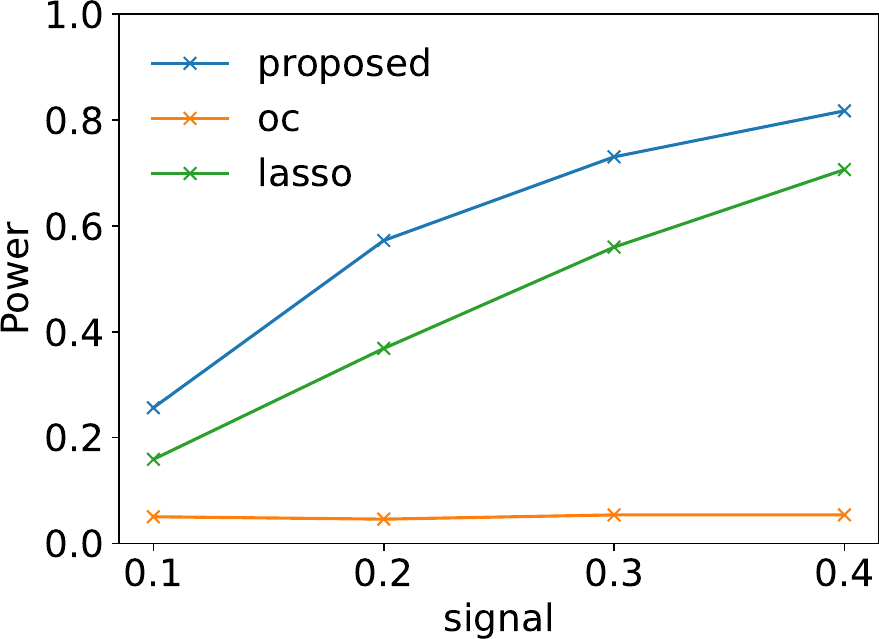}
        \subcaption{Independence}
    \end{minipage}
    \begin{minipage}[b]{0.49\linewidth}
        \centering
        \includegraphics[width=1.0\linewidth]{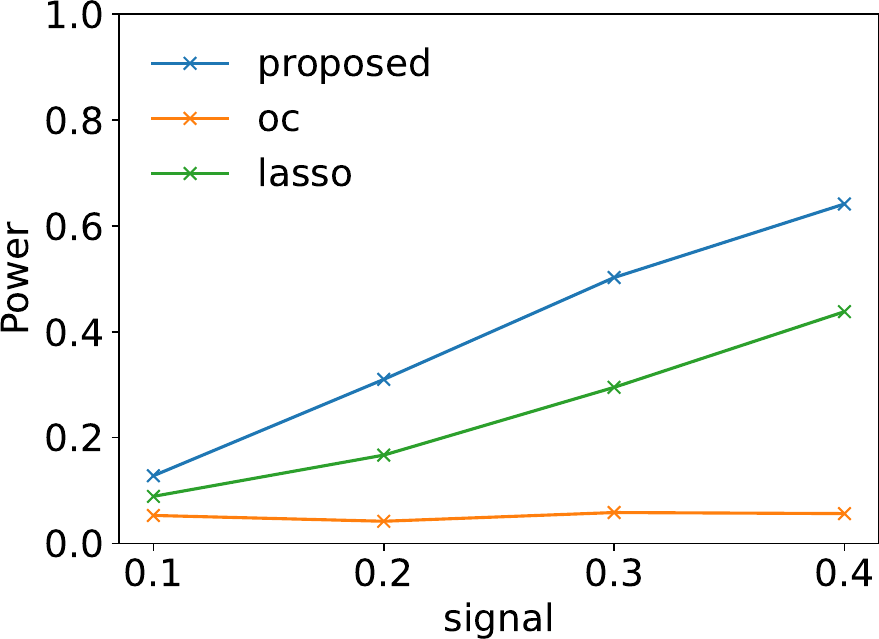}
        \subcaption{Correlation}
    \end{minipage}
    \caption{Power (linear-trend-shift)}
    \label{fig:tpr-linear-trend-shift}
    % \end{minipage}
\end{figure}

\begin{figure}[htbp]
    \centering
    \includegraphics[width=0.8\linewidth]{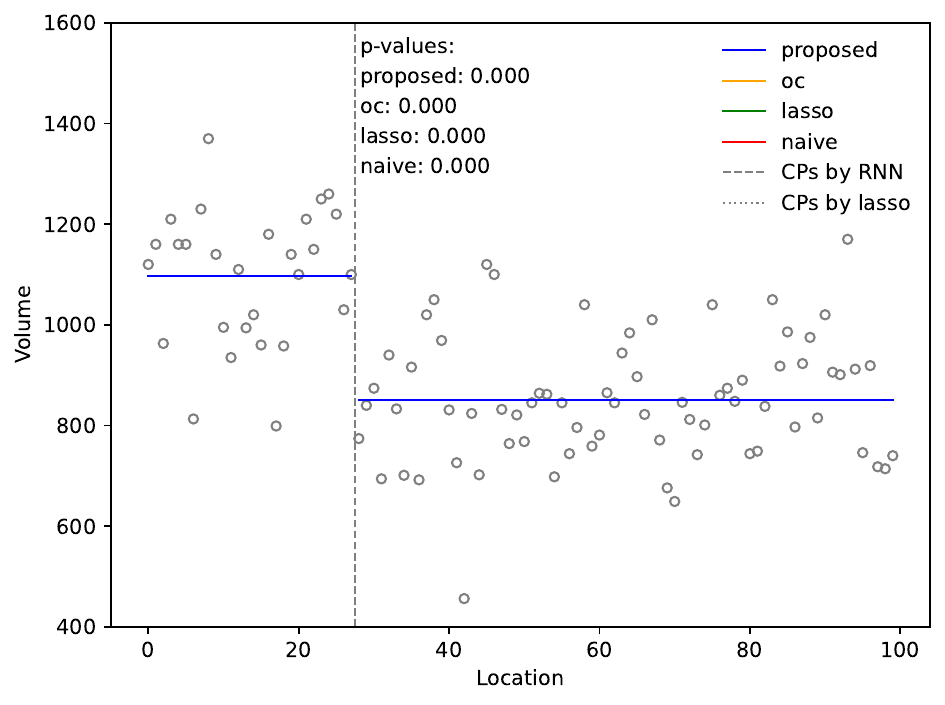}
    \caption{The Nile Dataset}
    \label{fig:nile}
\end{figure}

\begin{figure}[htbp]
    \begin{minipage}[b]{\linewidth}
        \centering
        \includegraphics[width=0.8\linewidth]{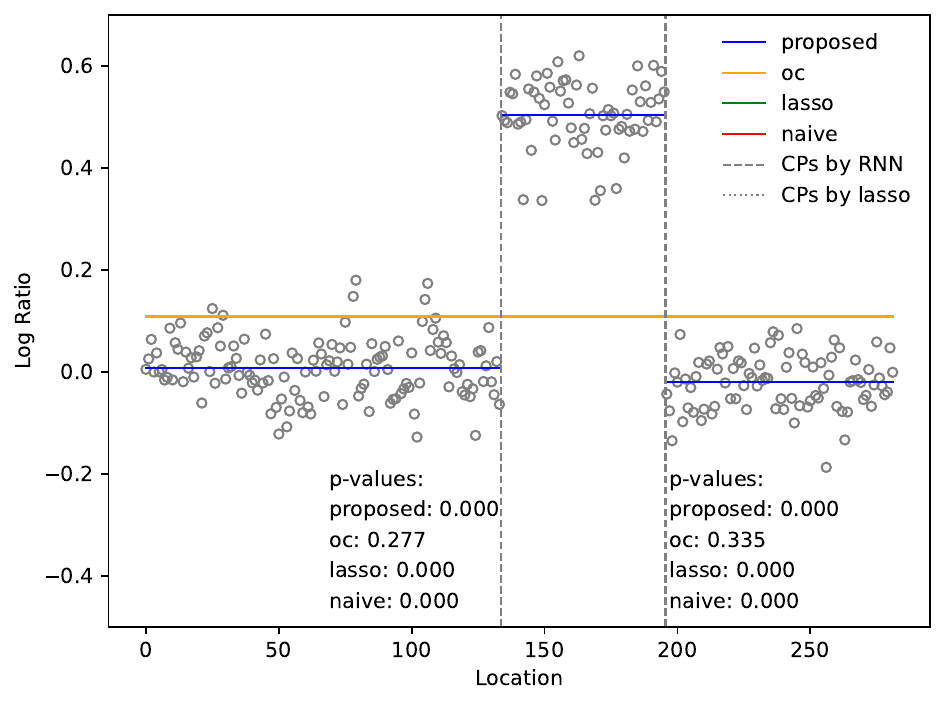}
        \subcaption{Cell line GM03576: Chromosomes 1,2,3.}
        \centering
        \includegraphics[width=0.8\linewidth]{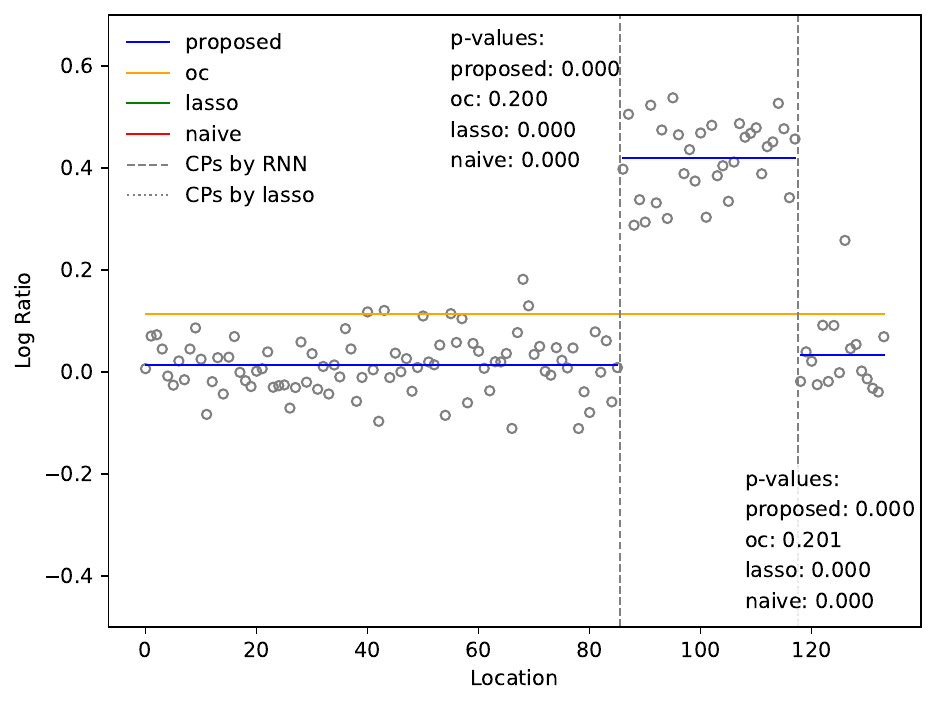}
        \subcaption{Cell line GM03576: Chromosomes 20,21,22.}
    \end{minipage}
    \caption{The Array CGH Dataset (1/2)}
    \label{fig:array_cgh_1}
\end{figure}

\begin{figure}[htbp]
    \begin{minipage}[b]{\linewidth}
        \centering
        \includegraphics[width=0.8\linewidth]{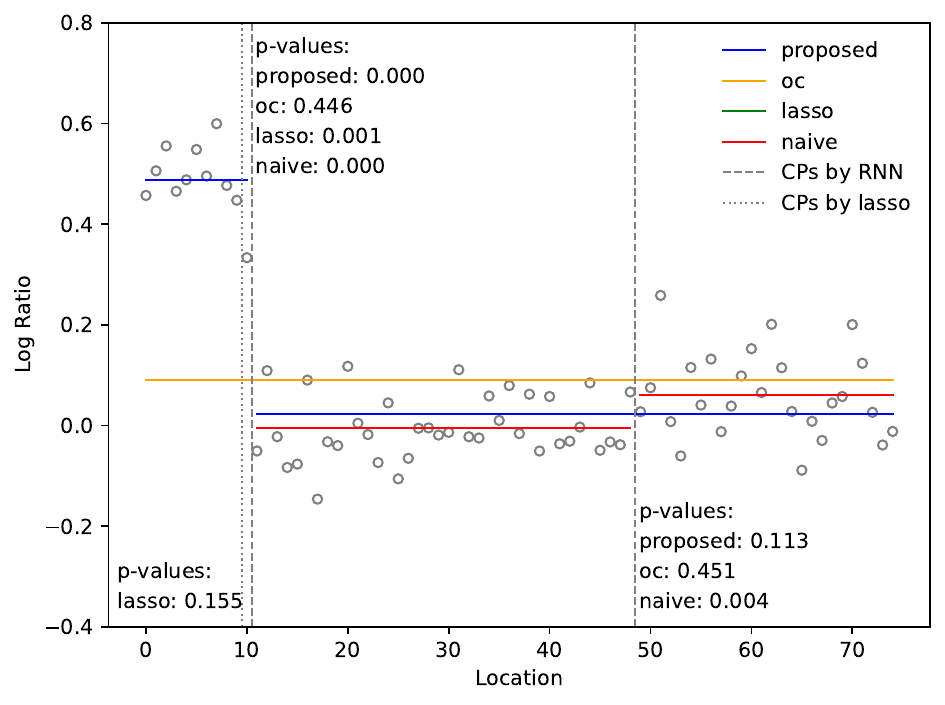}
        \subcaption{Cell line GM01750: Chromosomes 14.}
        \label{fig:array_cgh_2_a}
        \centering
        \includegraphics[width=0.8\linewidth]{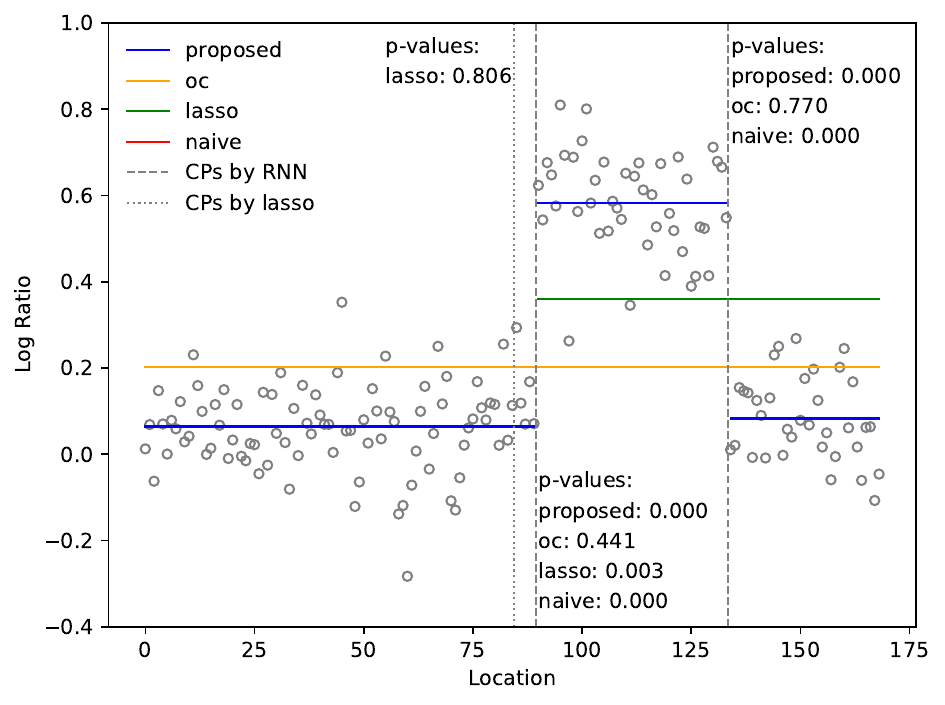}
        \subcaption{Cell line GM00143: Chromosomes 17,18,19.}
        \label{fig:array_cgh_2_b}
    \end{minipage}
    \caption{The Array CGH Dataset (2/2)}
    \label{fig:array_cgh_2}
\end{figure}

\begin{figure}
    \centering
    \includegraphics[width=0.8\linewidth]{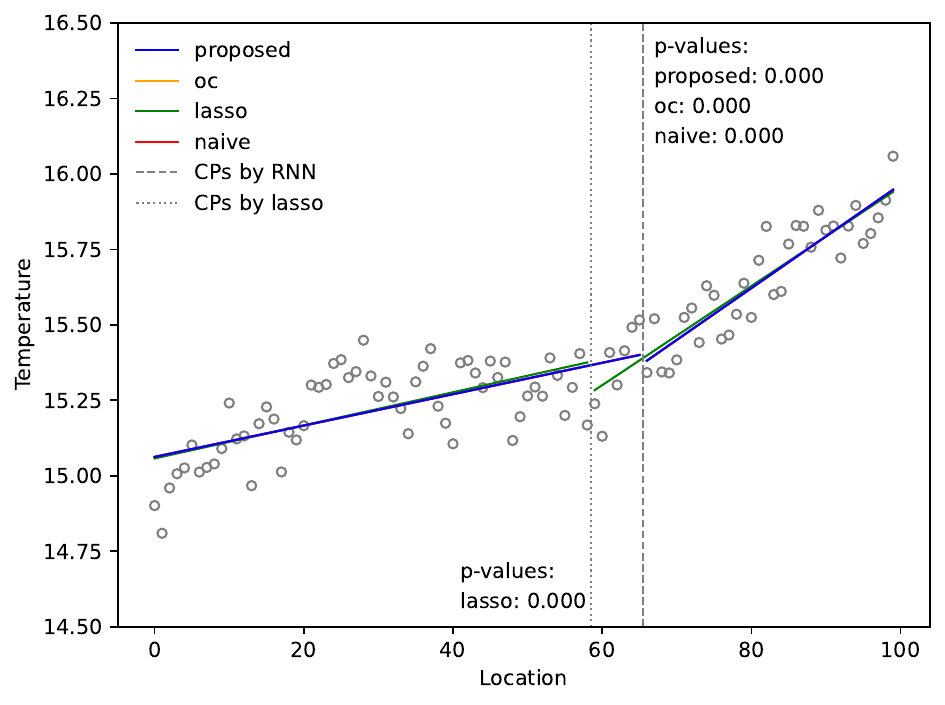}
    \caption{The Earth Surface Temperature Dataset}
    \label{fig:earth_surface_temperature}
\end{figure}

\begin{figure}[htbp]
    \begin{minipage}[b]{\linewidth}
        \centering
        \includegraphics[width=0.8\linewidth]{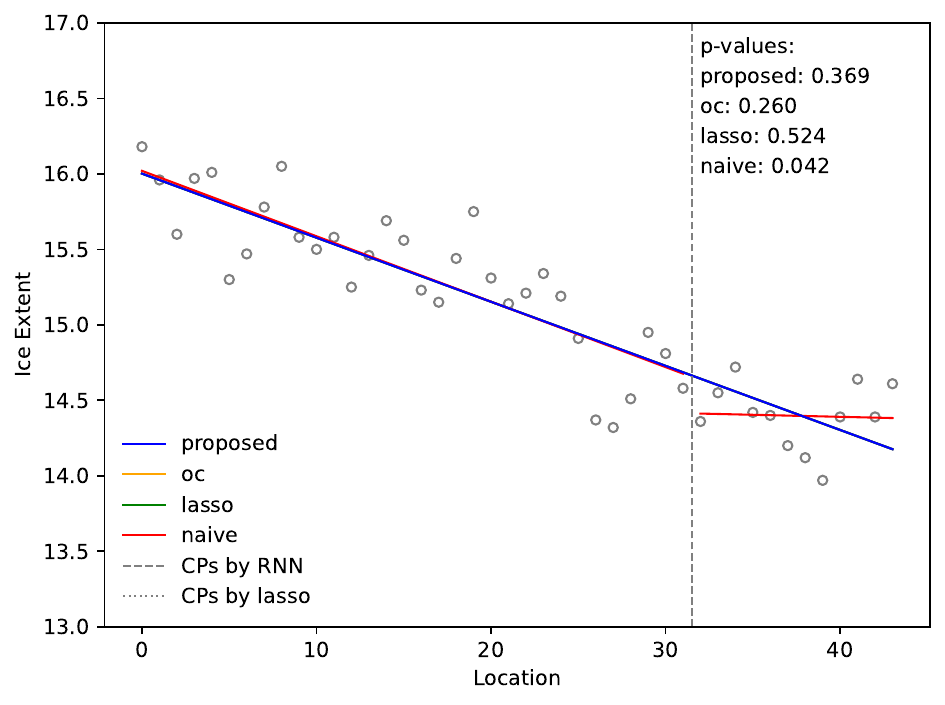}
        \subcaption{Arctic: February}
        \label{fig:ice_extent_1_a}
        \centering
        \includegraphics[width=0.8\linewidth]{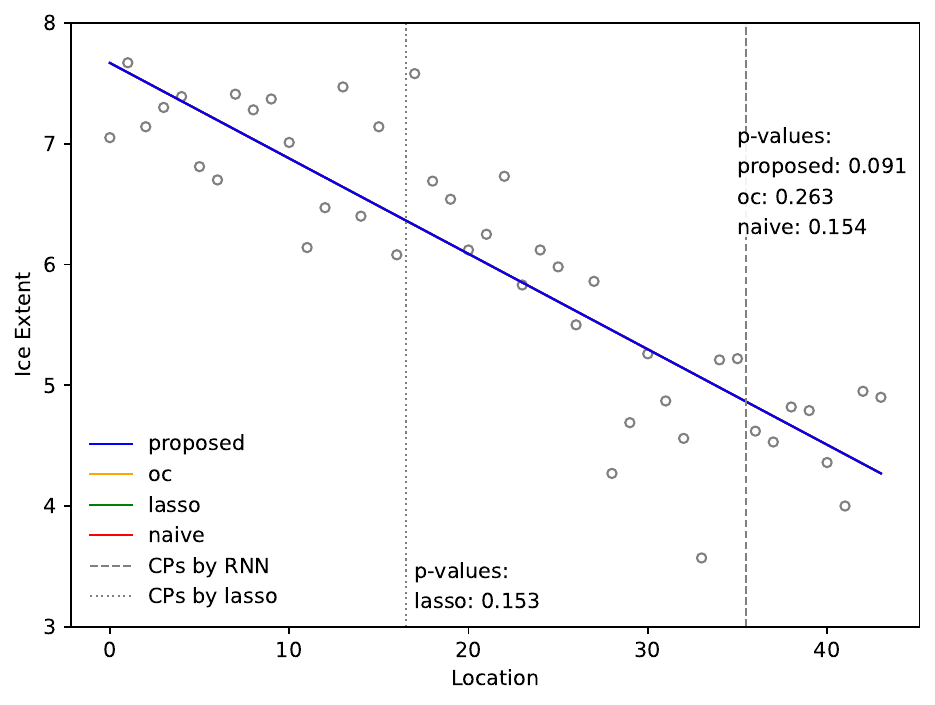}
        \subcaption{Arctic: September}
        \label{fig:ice_extent_1_b}
    \end{minipage}
    \caption{The Sea Ice Extent of Arctic and Antarctic Dataset (1/2)}
    \label{fig:ice_extent_1}
\end{figure}

\begin{figure}[htbp]
    \begin{minipage}[b]{\linewidth}
        \centering
        \includegraphics[width=0.8\linewidth]{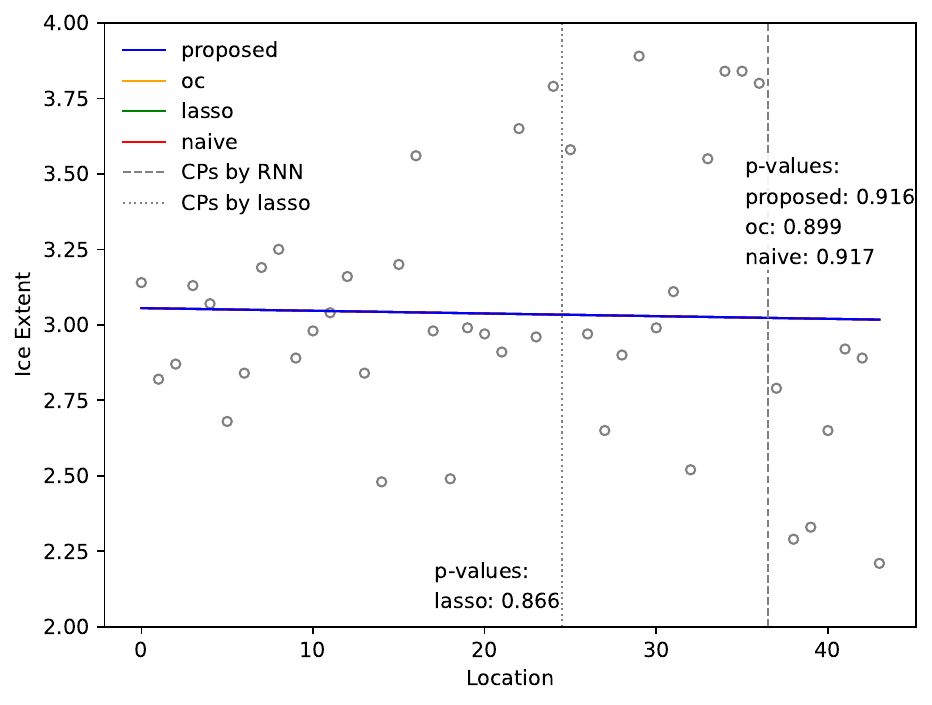}
        \subcaption{Antarctic: February}
        \centering
        \includegraphics[width=0.8\linewidth]{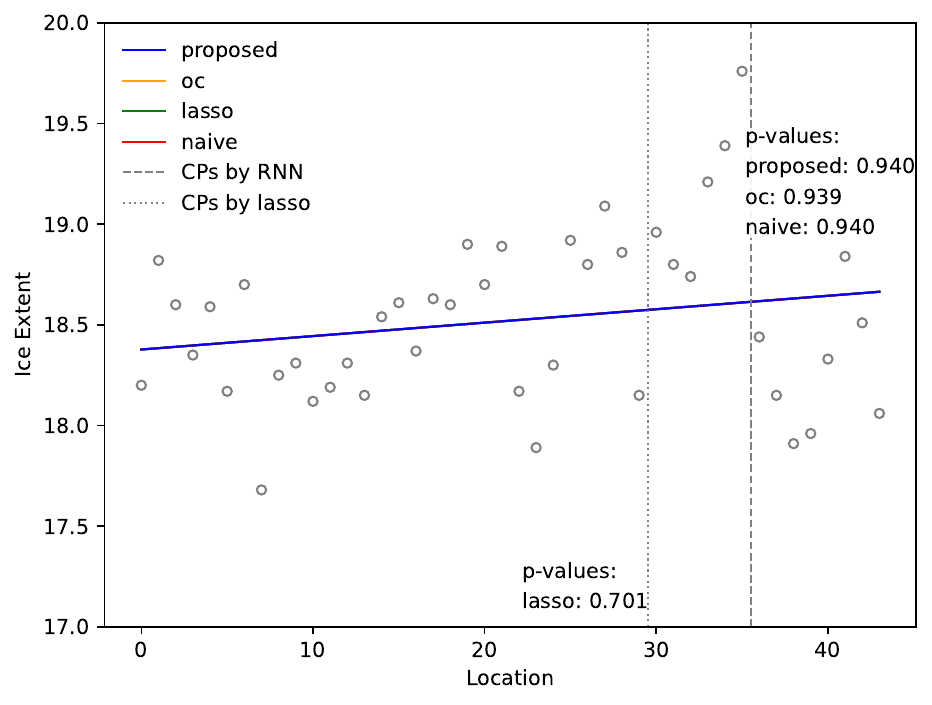}
        \subcaption{Antarctic: September}
    \end{minipage}
    \caption{The Sea Ice Extent of Arctic and Antarctic Dataset (2/2)}
    \label{fig:ice_extent_2}
\end{figure}

\paragraph{Methods for Comparison.}
We compared the proposed method (\texttt{proposed}) with three baselines: the generalized lasso path (\texttt{lasso}), the over-conditioning (\texttt{oc}), and the naive method (\texttt{naive}).
The \texttt{lasso} is a valid test for multiple detected CPs based on generalized lasso path~\citep{hyun2018exact}.
The \texttt{oc} is a simple application of polytope-based SI to RNN---we consider the \texttt{oc} to demonstrate the advantage of minimum conditioning.
The \texttt{naive} is the test without conditioning on the selection event.
The details of the baseline methods are shown in Appendix~\ref{app:experiment}.
\paragraph{Experimental Setup.}
In all experiments, we set $K=2$, $l=m=10$, $w=5$, and the significance level $\alpha=0.05$.
We considered two covariance matrices: $\Sigma=I_n\in\mathbb{R}^{n\times n}$ (independence) and $\Sigma=(0.5^{|i-j|})_{ij}\in\mathbb{R}^{n\times n}$ (correlation).
For the experiments to see the type I error rate, we generated 1,000 null sequences $\bm{x}=(x_1,\ldots,x_n)\sim\mathcal{N}(\bm{0}, \Sigma)$ with $n\in\{40, 60, 80, 100\}$ for both mean-shift and linear-trend-shift.
For each $n$, we run 1,000 trials.
To investigate the power, we generated sequences $\bm{x}=(x_1,\ldots,x_n)\sim\mathcal{N}(\bm{\mu}, \Sigma)$ with sample size $n=60$, in which mean vector $\bm{\mu}$ is defined as
\begin{equation}
    \mu_i =
    \begin{cases}
        0       & (i\in[1:20])  \\
        \Delta  & (i\in[21:40]) \\
        2\Delta & (i\in[41:60])
    \end{cases}
\end{equation}
for $\Delta\in\{1.0, 2.0, 3.0, 4.0\}$ for mean-shift, and
\begin{equation}
    \mu_i =
    \begin{cases}
        0            & (i\in[1:20])  \\
        (i-20)\Delta & (i\in[21:40]) \\
        20\Delta     & (i\in[41:60])
    \end{cases}
\end{equation}
for $\Delta\in\{0.1,0.2,0.3,0.4\}$ for linear-trend-shift.
In each case, we run 1,000 trials.
Since the tests are performed only when CPs are correctly detected, the power is defined as follows
\begin{equation}
    \text{Power (or Conditional Power)}
    =
    \frac
    {\text{\# correctly detected \& rejected}}
    {\text{\# correctly detected}}.
\end{equation}
Note that, since it is often difficult to accurately detect exact CPs in the presence of noise, many existing CP detection studies consider a detection to be correct if it is within $L$ positions of the true CP locations.
We follow the same convention and set $L=2$, i.e., we consider a detection to be correct if all detected CPs are within $\pm 2$ of the true CP locations.
\paragraph{Results.}
The results of type I error rate are shown in Figures~\ref{fig:fpr-mean-shift} and \ref{fig:fpr-linear-trend-shift}.
Three of the four methods \texttt{proposed}, \texttt{lasso} and \texttt{oc} successfully controlled the type I error rate under the significance level, whereas the \texttt{naive} could not.
Because the \texttt{naive} failed to control the type I error rate, we no longer considered its power.
The results of power are shown in Figures~\ref{fig:tpr-mean-shift} and \ref{fig:tpr-linear-trend-shift}.
We confirmed that the proposed method has the highest power.
The \texttt{oc} has the lowest power because it considers several extra conditions, which causes the loss of power.
\newpage
\paragraph{Real Data Experiments.}
We applied the proposed method and all the baseline methods to four real datasets.
We used the Nile dataset
\footnote{The information about this dataset is available at \url{https://stat.ethz.ch/R-manual/R-devel/library/datasets/html/Nile.html}}
and the Array CGH dataset~\citep{snijders2001assembly} for mean-shift setting,
while, we used the Earth Surface Temperature dataset
\footnote{This dataset is available at \url{https://www.kaggle.com/datasets/berkeleyearth/climate-change-earth-surface-temperature-data}}
and the monthly average sea ice extent of Arctic and Antarctic dataset~\citep{dataset_sea_ice_extent} for linear-trend-shift setting.
The results are shown in Figures~\ref{fig:nile}, \ref{fig:array_cgh_1}, \ref{fig:array_cgh_2}, \ref{fig:earth_surface_temperature}, \ref{fig:ice_extent_1}, and \ref{fig:ice_extent_2}, respectively.
Note that the \texttt{lasso} may detect different CPs than the other three methods \texttt{proposed}, \texttt{oc}, \texttt{naive} because CP detection algorithms are different.
The Nile dataset is the annual flow volume of the Nile river at Aswan from 1871 to 1970 (100 years).
In Figure~\ref{fig:nile}, we see that all methods detected a CP between 1988 and 1989.
This result is consistent with the widely noted fact~\citep{jung2017bayesian}.
This CP was found to be statistically significant in all methods.

The Array CGH dataset is used for detecting changes in expression levels across the genome.
The dataset with ground truth was provided in~\citep{snijders2001assembly}.
The results are shown in Figures~\ref{fig:array_cgh_1} and \ref{fig:array_cgh_2}.
All the results obtained by the proposed method are consistent with the ground truth~\citep{snijders2001assembly}.
However, no CP was found to be statistically significant by the \texttt{oc} due to its lack of power.
As shown in Figure~\ref{fig:array_cgh_2_b}, the \texttt{lasso} detected CP that differed from the proposed method and found it to be statistically insignificant.
The \texttt{naive} found the CP to be statistically significant as shown in Figure~\ref{fig:array_cgh_2_a}, which was a different result than the proposed method.

The Earth Surface Temperature dataset is preprocessed to the annual average temperatures at the land and oceans, from 1916 to 2015 (100 years).
In Figure~\ref{fig:earth_surface_temperature}, we see that the proposed method detected a CP between 1981 and 1982, while the \texttt{lasso} detected a CP between 1974 and 1975.
Both of these CPs were found to be statistically significant in all methods.
These results are consistent with the widely noted fact that global warming has been progressing more rapidly since the 1980s.

The sea ice extent of Arctic and Antarctic dataset is related to climate change in the Arctic and Antarctic.
As mentioned in~\citep{serreze2018arctic}, sea ice extent is the most common measure for assessing the feature of high-latitude oceans and it is defined as the area covered with an ice concentration of at least 15\%.
We used the data in February and September as an extreme case.
% because the Arctic has the maximum ice extent in February while the minimum occurs in September and the Antarctic does the opposite.
%
The results were shown in Figures~\ref{fig:ice_extent_1} and \ref{fig:ice_extent_2}.
As for previous studies, the decreasing trend of sea ice extent of the Arctic in September is noted to be one of the most important indicators of climate change in~\citep{serreze2018arctic}.
For the Antarctic, the sea ice extent was said to be stable and in a weak increasing trend in~\citep{comiso2017positive,serreze2018arctic}, but it was pointed out that this may have collapsed with the sharp drop in 2016 by~\citep{rintoul2018antarctica}.
However, when the data up to the year 2022 were re-examined, no significant CP was detected by three methods \texttt{proposed}, \texttt{oc}, and \texttt{lasso}.
Although the CPs detected by \texttt{proposed} and \text{lasso} are different, it is difficult to determine which is the better detection.
The \texttt{naive} found the CP to be statistically significant as shown in Figure~\ref{fig:ice_extent_1_a}, which was a different result than the proposed method.

\paragraph{Robustness of Type I Error rate Control.}
Additionally, we confirmed the robustness of the proposed method in terms of type I error rate control by applying our method to the case where variance is estimated from the same data.
To estimate the variance, we first performed CP detection algorithm to detect the
segments.
Since the estimated variance tends to be smaller than the true value, we computed the sample variance of each segment and set the maximum value as the variance to be used for the test.
We also considered the sequences with their noise following the five types of standardized non-gaussian distribution families (the details are shown in Appendix~\ref{app:experiment}).

In experiment for estimated variance, we generated 1,000 null sequences for $n\in\{40,60,80,100\}$ and tested the type I error rate for $\alpha = 0.01, 0.05, 0.1$.
To check robustness for non-gaussian noise, we first obtained a distribution such that the 1-Wasserstein distance from $\mathcal{N}(0,1)$ is $d$ in each distribution family, for $d\in\{0.03, 0.06, 0.09, 0.12, 0.15\}$
We then generated 1.000 sequences following each distribution with sample size $n=60$.
We confirmed that our method still maintains good performance on type I error rate control.
The results are shown in Appendix~\ref{app:experiment}.
\newpage
%
% \addtocounter{figure}{-1}
%
\newpage
\clearpage
%
% \addtocounter{figure}{-1}
%

\newpage
\section{Conclusion}
\label{sec:sec5}
In this paper, we introduce a novel method for conducting statistical inference on the change points (CPs) detected by RNNs.
Our approach marks a significant advancement as it is the first method capable of providing valid $p$-values for hypotheses driven by RNNs.
Through comprehensive experiments on both synthetic and real-world datasets, our results robustly support the theoretical results of our method, demonstrating its superior performance compared to existing approaches.
The potential future work could be extending our method to the case of multi-dimensional sequences and more sophisticated RNN architectures, such as Long Short-Term Memory (LSTM) and Gated Recurrent Unit (GRU).
The intricate gating mechanism in advanced RNNs poses challenges in characterizing selection events, making the development of an efficient approach a valuable future contribution.

\newpage
\subsection*{Acknowledgement}
This work was partially supported by MEXT KAKENHI (20H00601), JST CREST (JPMJCR21D3, JPMJCR22N2), JST Moonshot R\&D (JPMJMS2033-05), JST AIP Acceleration Research (JPMJCR21U2), NEDO (JPNP18002, JPNP20006) and RIKEN Center for Advanced Intelligence Project.

\clearpage
\appendix
\newpage
\section{Experimental Details}
\label{app:experiment}
\begin{figure}[htbp]
    \centering
    \includegraphics[width=0.6\linewidth]{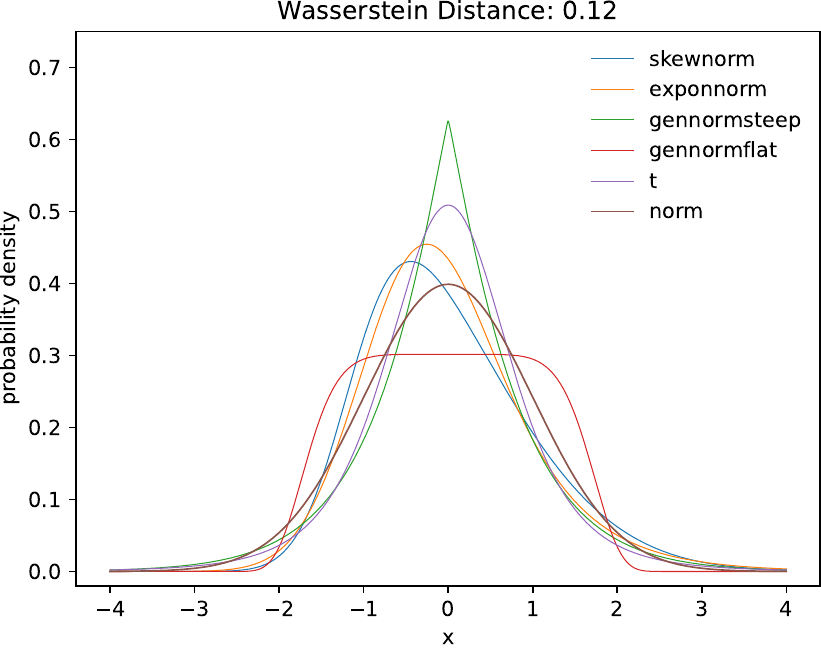}
    \caption{Demonstration of non-gaussian distributions}
    \label{fig:non_gaussian}
\end{figure}

\begin{figure}[htbp]
    % \begin{minipage}[b]{0.49\linewidth}
    \begin{minipage}[b]{0.49\linewidth}
        \centering
        \includegraphics[width=1.0\linewidth]{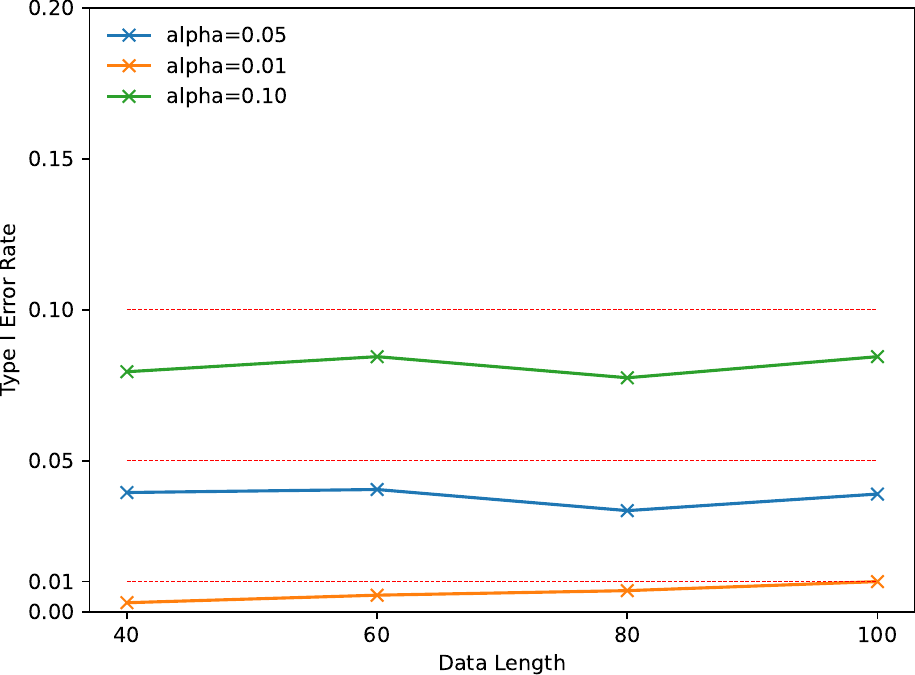}
        \subcaption{mean-shift}
    \end{minipage}
    \begin{minipage}[b]{0.49\linewidth}
        \centering
        \includegraphics[width=1.0\linewidth]{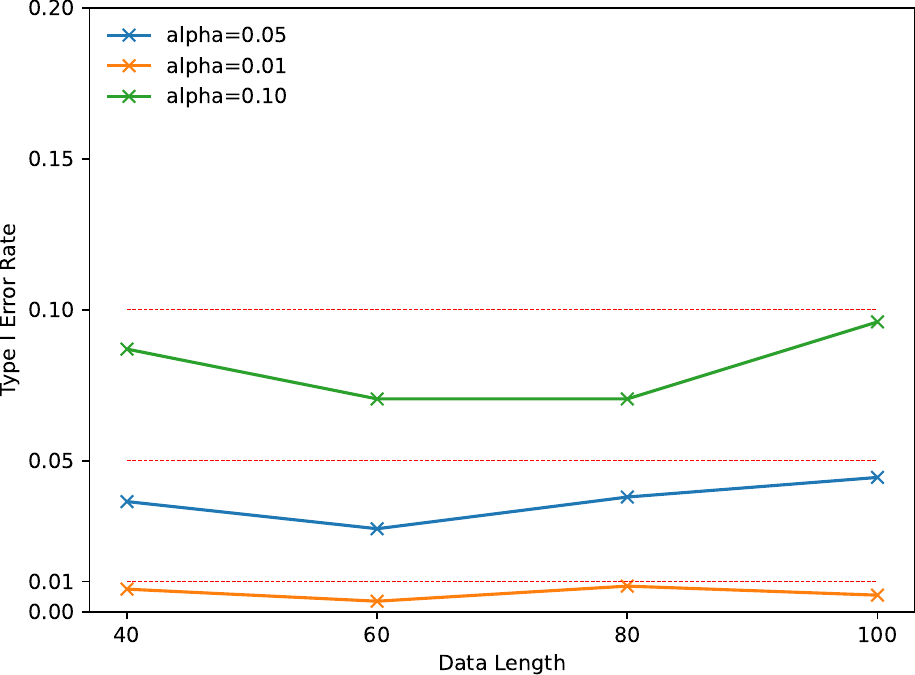}
        \subcaption{linear-trend-shift}
    \end{minipage}
    \caption{Robustness of type I error rate control for estimated variance}
    \label{fig:robustness_estimated}
\end{figure}

\begin{figure}[htbp]
    \begin{minipage}[b]{0.49\linewidth}
        \centering
        \includegraphics[width=1.0\linewidth]{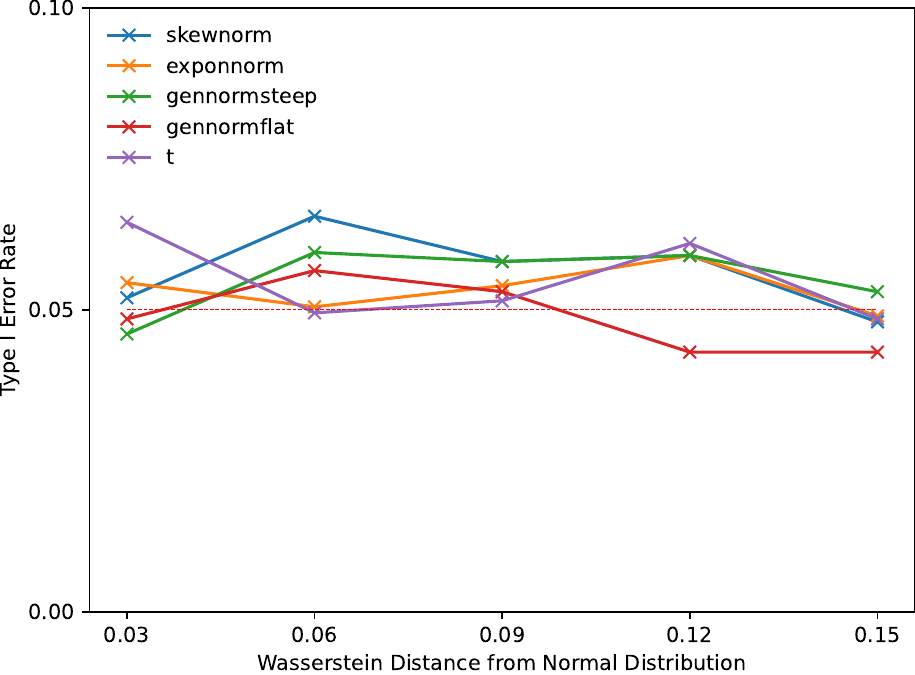}
        \subcaption{mean-shift}
    \end{minipage}
    \begin{minipage}[b]{0.49\linewidth}
        \centering
        \includegraphics[width=1.0\linewidth]{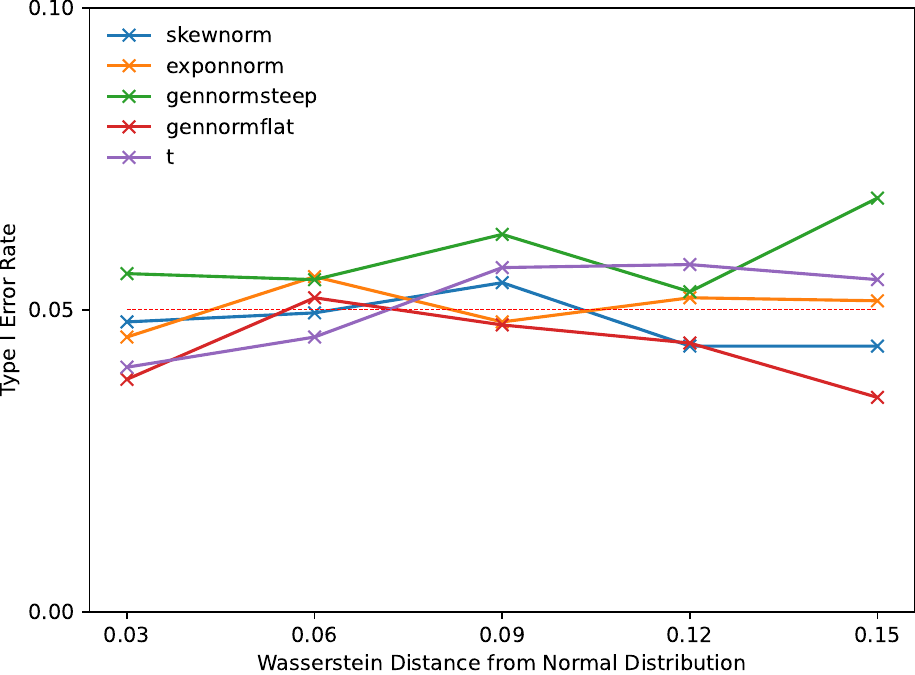}
        \subcaption{linear-trend-shift}
    \end{minipage}
    \caption{Robustness of type I error rate control for non-gaussian noise}
    \label{fig:robustness_non_gaussian}
\end{figure}

\paragraph{Methods for comparison.}
We compared our proposed method with the following baselines:
\begin{itemize}
    \item \texttt{lasso}: This is a method to validly test the CPs detected by a generalized lasso path by applying SI, and was proposed in~\citep{hyun2018exact}.
    \item \texttt{oc}: This is a method of additionally conditioning on the $\mathcal{S}$, i.e., following the notation in \eqref{eq:selective_p_from_region}, we compute the $p$-values as
          \begin{equation*}
              p^\mathrm{oc}_k = \mathbb{P}_{\mathrm{H}_{0,k}}
              \left(
              \left|Z\right| \geq \left|Z^\mathrm{obs}\right|\mid
              Z\in\mathcal{Z}^\mathrm{oc}(Z^\mathrm{obs})
              \right).
          \end{equation*}
    \item \texttt{naive}: This is a naive testing method without conditioning, i.e., we compute the $p$-values as
          \begin{equation*}
              p^\mathrm{naive}_k = \mathbb{P}_{\mathrm{H}_{0,k}}
              \left(
              \left|Z\right| \geq \left|Z^\mathrm{obs}\right|
              \right).
          \end{equation*}
\end{itemize}
\paragraph{Non-Gaussian Distribution Families}
We considered the following non-gaussian distribution families:
\begin{itemize}
    \item \texttt{skewnorm}: Skew normal distribution family.
    \item \texttt{exponnorm}: Exponentially modified normal distribution family.
    \item \texttt{gennormsteep}: Generalized normal distribution family (limit the shape parameter $\beta$ to be steeper than the normal distribution, i.e., $\beta < 2$).
    \item \texttt{gennormflat}: Generalized normal distribution family (limit the shape parameter $\beta$ to be flatter than the normal distribution, i.e., $\beta > 2$).
    \item \texttt{t}: Student's t distribution family.
\end{itemize}
Note that these are all distribution families, including the normal distribution, and are standardized in the experiment.
We demonstrate the probability density functions for distributions from each distribution family such that the 1-Wasserstein distance from $\mathcal{N}(0,1)$ is $0.12$, in Figure~\ref{fig:non_gaussian}
\paragraph{Robustness of Type I Error Rate Control.}
The results of estimated variance are shown in Figure~\ref{fig:robustness_estimated} and our proposed method can properly control the type I error rate.
The results of non-gaussian noise are shown in Figure~\ref{fig:robustness_non_gaussian}.
In case of mean-shift, except for \texttt{gennormflat}, the type I error rate is just a bit higher than the significance level.
For linear-trend-shift, in case of \texttt{skewnorm}, \texttt{exponnorm} and \texttt{gennormflat}, our proposed method still can properly control the type I error rate.
But, in case of \texttt{gennormsteep} and \texttt{t}, the type I error rate is just a bit higher than the significance level.
\newpage
\bibliographystyle{plainnat}
\bibliography{ref}

\end{document}